\documentclass[10pt,letterpaper,twoside,twocolumn,journal,final]{IEEEtran}
\usepackage{subfigure}
\usepackage[ruled,vlined,linesnumbered]{algorithm2e}
\usepackage[abs]{overpic}
\usepackage{graphicx}
\usepackage{textcomp}
\usepackage{gnuplot-lua-tikz}
\usepackage{pgfplots}
\newcommand{\figwidth}{1.3in}
\newcommand{\figscale}{0.4}
\newcommand{\figfontsize}{\small}
\newcommand{\smallfigwidth}{1.0in}
\newcommand{\zsmallfigwidth}{1.1in}
\newcommand{\graphwidth}{3.5in}
\usetikzlibrary{arrows,decorations.pathmorphing,decorations.markings,backgrounds,positioning,fit,petri}
\graphicspath{{figs/}}
\begin{document}
\title{Toward the Coevolution of Novel Vertical-Axis \\ Wind Turbines}
\author{Richard~J.~Preen and Larry~Bull%
	\thanks{The final version of this paper is published in IEEE Transactions on Evolutionary Computation, doi:10.1109/TEVC.2014.2316199}%
	\thanks{This work was supported in part by the UK Leverhulme Trust under Grant RPG-2013-344.}%
\thanks{The authors are with the Department of Computer Science and Creative Technologies, University of the West of England, Bristol BS16 1QY, U.K. (e-mail: richard2.preen@uwe.ac.uk, larry.bull@uwe.ac.uk).}}
\markboth{}
{PREEN \MakeLowercase{and} BULL:~TOWARD THE COEVOLUTION OF NOVEL VERTICAL-AXIS WIND TURBINES}
\maketitle
\begin{abstract}
The production of renewable and sustainable energy is one of the most important challenges currently facing mankind. Wind has made an increasing contribution to the world's energy supply mix, but still remains a long way from reaching its full potential. In this paper, we investigate the use of artificial evolution to design vertical-axis wind turbine prototypes that are physically instantiated and evaluated under fan generated wind conditions. Initially a conventional evolutionary algorithm is used to explore the design space of a single wind turbine and later a cooperative coevolutionary algorithm is used to explore the design space of an array of wind turbines. Artificial neural networks are used throughout as surrogate models to assist learning and found to reduce the number of fabrications required to reach a higher aerodynamic efficiency. Unlike in other approaches, such as computational fluid dynamics simulations, no mathematical formulations are used and no model assumptions are made.

\end{abstract}
\IEEEpeerreviewmaketitle%
\section{Introduction}
\IEEEPARstart{I}{n recent} years, wind has made an increasing contribution to the world's energy supply mix. However, there is still much to be done in all areas of the technology for it to reach its full potential. Currently, horizontal-axis wind turbines (HAWTs) are the most commonly used form. However, ``modern wind farms comprised of HAWTs require significant land resources to separate each wind turbine from the adjacent turbine wakes. This aerodynamic constraint limits the amount of power that can be extracted from a given wind farm footprint. The resulting inefficiency of HAWT farms is currently compensated by using taller wind turbines to access greater wind resources at high altitudes, but this solution comes at the expense of higher engineering costs and greater visual, acoustic, radar and environmental impact''~\cite{Dabiri:2011}. This has forced wind energy systems away from high energy demand population centres and towards remote locations with higher distribution costs. In contrast, vertical-axis wind turbines (VAWTs) do not need to be oriented to wind direction and can be positioned closely together, potentially resulting in much higher efficiency. VAWT can also be easier to manufacture, may scale more easily, are typically inherently light-weight with little or no noise pollution, and are more able to tolerate extreme weather conditions~\cite{Eriksson:2008}. However, their design space is complex and relatively unexplored. Generally, two classes of design are currently under investigation and exploitation: the Savonius~\cite{Savonius:1930}, which has blades attached directly upon the central axis structure; and the Darrieus~\cite{Darrieus:1931}, where the blades---either straight or curved---are positioned predominantly away from the central structure. Hybrids also exist.

The majority of blade design optimisation is performed through the use of computational fluid dynamics (CFD) simulations, typically described with 3D Navier-Stokes equations~\cite{Anderson:1995}. However, 3D CFD simulations are computationally expensive, with a single calculation taking hours on a high-performance computer, making their use with an iterative search approach difficult~\cite{Graning:2007}. Moreover, assumptions need to be made, e.g.,\ regarding turbulence or pressure distributions, which can significantly affect accuracy. Previous evolutionary studies have been undertaken with types of CFD to optimise the blade profile for both HAWT~\cite{Hampsey:2002} and VAWT~\cite{Carrigan:2012} to varying degrees of success/realism.

Evolutionary algorithms (EAs) have been used to design 3D physical objects, such as furniture~\cite{BentleyWakefield:1995}, aircraft engine blades~\cite{LeeHajela:1996} and wings~\cite{OngKeane:2004}. Notably, Lohn {\it et~al.\/}~\cite{Lohn:2008} evolved and manufactured an {X}-band satellite antenna for {NASA}'s {ST5} spacecraft, representing the world's first artificially evolved hardware in space. Significantly, the antenna's performance was similar to a design hand-produced by an antenna-contractor. Most of these approaches, however, have used simulations to provide the fitness scores of the evolved designs. 

The evaluation of physical artifacts for fitness determination can be traced back to the origins of evolutionary computation; for example, the first evolution strategies (ESs) were used to design jet nozzles with a string of section diameters, which were then machined and tested for fitness~\cite{Rechenberg:1971}. Other well-known examples include robot controller design~\cite{Nolfi:1992}, electronic circuit design using programmable hardware~\cite{Thompson:1998}, product design via human provided fitness values~\cite{Herdy:1996}, chemical systems~\cite{Theis:2007}, and unconventional computers~\cite{HardingMiller:2004}. Evolution in hardware has the potential to benefit from access to a richer environment where it can exploit subtle interactions that can be utilised in unexpected ways. For example, the EA used by Thompson~\cite{Thompson:1998} to work with field-programmable gate array circuits used some subtle physical properties of the system to solve problems where the properties used are still not understood. Humans can be prevented from designing systems that exploit these subtle and complex physical characteristics through their lack of knowledge, however this does not prevent exploitation through artificial evolution. There is thus a real possibility that evolution in hardware may allow the discovery of new physical effects, which can be harnessed for computation/optimisation~\cite{MillerDowning:2002}.  

Moreover, the advent of high quality, low-cost, additive rapid fabrication technology---known as 3D printing---means it is now possible to fabricate a wide range of prototype designs quickly and cheaply. 3D printers are now capable of printing an ever growing array of different materials, including food (e.g.,\ chocolate~\cite{Hao:2009} and meat~\cite{Lipton:2010} for culinary design), sugar (e.g.,\ to help create synthetic livers~\cite{Miller:2012}), chemicals (e.g.,\ for custom drug design~\cite{Cronin:2012}), cells (e.g.,\ for functional blood vessels~\cite{Jakeb:2008} and artificial cartilage~\cite{Xu:2013}), plastic (e.g.,\ Southampton University laser sintered aircraft), thermoplastic (e.g.,\ for electronic sensors~\cite{Leigh:2012}), titanium (e.g.,\ for prosthetics such as the synthetic mandible developed by the University of Hasselt and transplanted into an 83-year old woman), and liquid metal (e.g.,\ for stretchable electronics~\cite{Ladd:2013}). One potential benefit of the technology is the ability to perform fabrication directly in the target environment; for example, Cohen {\it et~al.\/}~\cite{Cohen:2010} recently used a 3D printer to perform a minimally invasive repair of the cartilage and bone of a calf femur {\it in situ}. Lipson and Pollack~\cite{LipsonPollack:2000} were the first to exploit the emerging technology in conjunction with an EA using a simulation of the mechanics and control, ultimately printing mobile robots with embodied neural network controllers.

The use of surrogate models (also known as meta-models or response surface models) to reduce the number of candidate solution evaluations when they are computationally expensive or difficult to obtain/formulate has been developed as evolutionary computation has been applied to more complex domains, e.g.,\ in systems where a human user is involved~\cite{Bull:1997a}. This is typically achieved through the construction of models of the problem space via direct sampling---the use of approximations is an established approach in the wider field of optimisation~\cite{Dunham:1963}. That is, the evolutionary process uses one or more models to provide the (approximate) utility of candidate solutions, thereby reducing the number of real evaluations during iterations. Initially, all candidate solutions must be evaluated directly on the task to provide rudimentary training data for the modelling, e.g.,\ by neural networks. Periodically, high utility solutions suggested by the model optimisation are then evaluated by the real system. The training data for the model is then augmented with these and the model(s) updated. Over time, as the quality of the model(s) improves, the need to perform real evaluations/fabrications reduces.

In this paper, we present results from a pilot study of surrogate-assisted EAs (SGAs) used to design VAWTs wherein prototypes are evaluated under fan generated wind conditions after being physically instantiated by a 3D printer. That is, unlike other approaches, no mathematical formulations are used and no model assumptions are made. Initially, artificial evolution is used to explore the design space of a single isolated VAWT and subsequently a coevolutionary genetic algorithm (CGA) is applied to explore the design space of an array of 2 closely positioned VAWTs. Both conventional EA and SGA versions are examined. To date, no prior work has explored the use of 3D printing within surrogate-assisted embodied evolution. 
\section{Related Work}
The evolution of geometric models to design arbitrary 3D morphologies has been widely explored. Early examples include Watabe and Okino's lattice deformation approach~\cite{WatabeOkino:1993} and McGuire's sequences of polygonal operators~\cite{McGuire:1993}. Sims~\cite{Sims:1994} evolved the morphology and behaviour of virtual creatures that competed in simulated 3D worlds with a directed graph encoding. Bentley~\cite{Bentley:1996} investigated the creation of 3D solid objects via the evolution of both fixed and variable length direct encodings. The objects evolved included tables, heatsinks, penta-prisms, boat hulls, aerodynamic cars, as well as hospital department layouts. Eggenberger~\cite{Eggenberger:1997} evolved 3D multicellular organisms with differential gene expression. Jacob and Nazir~\cite{JacobNazir:2002} evolved polyhedral objects with a set of functions to manipulate the designs by adding stellating effects, shrinking, truncating, and indenting polygonal shapes. More recently, Jacob and Hushlak~\cite{JacobHushlak:2007} used an interactive evolutionary approach with L-systems~\cite{PrusinkiewiczLindenmayer:1990} to create virtual sculptures and furniture designs.

EAs have also been applied to aircraft wing design~\cite{OngKeane:2004} including aerodynamic transonic aerofoils~\cite{HaciogluOzkol:2003,QuagliarellaCioppa:1995}, and multidisciplinary blade design~\cite{HajelaLee:1995}. Few evolved designs, however, have been manufactured into physical objects. Conventionally evolved designs tend to be purely descriptive, specifying what to build but not how it should be built. Thus, there is always an inherent risk of evolving interesting yet unbuildable objects. Moreover, high-fidelity simulations are required to ensure that little difference is observed once the virtual design is physically manifested. In highly complex design domains, such as dynamic objects, the difference between simulation and reality is too large to manufacture designs evolved under a simulator, and in others the simulations are extremely computationally expensive.

Funes and Pollack~\cite{FunesPollack:1998} performed one of the earliest attempts to physically instantiate evolved 3D designs by placing physical {LEGO} bricks according to the schematics of the evolved individuals. A direct encoding of the physical locations of the bricks was used and the fitness was scored using a simulator which predicted the stability of the composed structures. Additionally, Hornby and Pollack~\cite{HornbyPollack:2001} used L-systems to evolve furniture designs, which were then manufactured by a 3D printer. They found the generative encoding of L-systems produced designs faster and with higher fitness than a non-generative system. Generative systems are known to produce more compact encodings of solutions and thereby greater scalability than direct approaches~\cite{Schoenauer:1996}.

Compositional pattern producing networks~\cite{Stanley:2007} have recently been used to evolve 3D objects which were ultimately fabricated on a 3D printer~\cite{AuerbachBongard:2010a,AuerbachBongard:2010b,CluneLipson:2011}. Both interactive and target-based approaches were explored.

Recently, Rieffel and Sayles~\cite{RieffelSayles:2010} evolved circular two-dimensional shapes where each design was fabricated on a 3D printer before assigning fitness values. Interactive evolution was undertaken wherein the fitness for each printed shape was scored subjectively. Each individual's genotype consisted of twenty linear instructions which directed the printer to perform discrete movements and extrude the material. As a consequence of performing embodied fitness evaluations the system as a whole can exhibit epigenetic traits, where phenotypic characteristics arise from the mechanics of assembly. One such example was found when selecting shapes that most closely resembled the letter `A'. In certain individuals, the cross of the pattern was produced from the print head dragging a thread of material as it moved between different print regions and was not explicitly instructed to do so by the genotype.

The application of EAs can be prohibitive when the evaluations are computationally expensive or an explicit mathematical fitness function is unavailable. Whilst the speed and cost of rapid-prototyping continues to improve, fabricating an evolved design before fitness can be assigned remains an expensive task when potentially thousands of evaluations are required (e.g.,\ 10{\it mins\/} print time for each very simple individual in~\cite{RieffelSayles:2010}). However, given a sample $\mathcal{D}$ of evaluated individuals $N$, a surrogate model, $y=f(\vec{x})$, can be constructed, where $\vec{x}$ is the genotype, and $y$, fitness, in order to compute the fitness of an unseen data point $\vec{x} \notin \mathcal{D}$. The use of surrogate models has been shown to reduce the convergence time in evolutionary computation and multiobjective optimisation; see~\cite{Jin:2005,Jin:2011,Lim:2010} for recent reviews. Alternative methods, such as fitness inheritance~\cite{Smith:1995}, fitness imitation~\cite{KimCho:2001}, and fitness assignment~\cite{SalamiHendtlass:2003} can also be used.

Typically, a set of evaluated genotypes and their real fitness scores are used to perform the supervised training of an {MLP}-based artificial neural network; e.g.,~\cite{Bull:1999}. However, other approaches have been explored, e.g.,\ kriging~\cite{Ratle:2001}, clustering~\cite{KimCho:2001}, support vector regression~\cite{Yun:2009}, radial-basis functions~\cite{Ong:2006}, and sequential parameter optimisation~\cite{Bartz-Beielstein:2006}. The surrogate model is subsequently used to compute estimated fitness values for the EA to utilise. The model must be periodically retrained with new individuals under a controlled evolutionary approach to prevent convergence on local optima. Retraining can be performed by taking either an individual or generational approach~\cite{Jin:2000}. In the individual approach, $n$ number of individuals in the population are chosen and evaluated with the real fitness function each generation. In the generational approach, the entire population is evaluated on the real fitness function each $n$-th generation. Resampling methods and surrogate model validation remain an important and ongoing area of research, enabling the comparison and optimisation of models~\cite{Bischl:2012}. Both global modelling and local modelling using trust regions (e.g.,\ samples within a certain euclidean distance) are popular approaches.  

Surrogate assisted EAs that use CFD analysis for fitness determination have previously been used to design turbine blades, finding interesting solutions with reduced computational time~\cite{Song:2002}. Jin {\it et~al.\/}~\cite{Jin:2002} explored an ES with CFD analysis to minimise the pressure loss of a turbine blade while maintaining a certain outflow angle. The blade representation used consisted of a series of B-spline control points. The population was initialised with a given blade and 2 neural networks were used to approximate the pressure loss and outflow angle, finding faster convergence than without the surrogate models. Gr\"{a}ning {\it et~al.\/}~\cite{Graning:2007} used an ES with covariance matrix adaptation to minimise the pressure loss of a blade using 3D CFD simulations. The ES was augmented by a neural network surrogate model and used a pre-selection resampling approach (where offspring are only generated from individuals evaluated on the real fitness function), however significant improvement over a simple ES was not found.

The surrogate assisted evolution of aerofoil geometries (a type of blade) has been widely explored for use with aircraft design. Some examples include, Giotis and Giannakoglou~\cite{GiotisGiannakoglou:1999} who used multiple output neural networks as surrogate models for multiobjective aerofoil optimisation. Emmerich and Naujoks~\cite{EmmerichNaujoks:2004} and Kumano {\it et al.\/}~\cite{Kumano:2006} used kriging to provide approximations for multiobjective aerofoil design. In addition, Zhou {\it et~al.\/}~\cite{Zhou:2007} evolved aerodynamic aerofoil geometries with a representation consisting of Hicks-Henne bump function parameters. The EA was assisted by both a global and local surrogate model. Significantly, these approaches use simulations to evaluate candidate solutions and typically consider only two-dimensional blades (due to the cost of CFD analysis).
\section{Single VAWT Rotation-Based Evolution}
\label{sec:single}
Here we initially explore the evolution of a single VAWT with a non-varying $z$-axis, more representational flexibility is then introduced by enabling $z$ variability, and subsequently the coevolution of an array of 2 closely positioned $z$-varying VAWT prototypes is examined. Both conventional EA and surrogate-assisted approaches are used throughout.

A vector of 10 integers is used as a simple and compact encoding of the $x$-$y$-axis of a prototype VAWT. Each allele thus controls 1/10th of a single $z$-layer. A workspace (maximum object size) of $30\times30\times30mm$ is used so that the instantiated prototype is small enough for timely production ($\sim30mins$) and with low material cost, yet large enough to be sufficient for fitness evaluation. The workspace has a resolution of $100\times100\times100$ voxels. A central platform is constructed for each individual to enable the object to be placed on to the evaluation equipment. The platform consists of a square torus, 1 voxel in width and with a centre of $14\times14$ empty voxels that are duplicated for each $z$ layer, thus creating a hollow tube that is $3mm$ in diameter.

\begin{figure}[t]
	\centering
		% (a)
		\subfigure[]{\label{fig:d1}
			\begin{tikzpicture}[scale=\figscale, font=\figfontsize] 
				% grid
				\draw [thick, draw=black, fill=white] (0,0) grid (4,10) rectangle (0,0);
				\draw [thick, draw=black, fill=black] (0,0) rectangle (1,5);
				% x axis label
				\foreach \i in {1,...,4} {
					\draw (\i,1) -- (\i,10) node [below] at (\i-0.5,0) {$\i$};
				}
				\node[] at (2,-1.5) {gene no.};
				% y axis label
				\foreach \i in {1,...,10} {
					\draw (1,\i) -- (4,\i) node [left] at (0,\i) {$\i$};
				}
				\node[rotate=90] at (-1.5,5) {gene value};
			\end{tikzpicture}
		}%
		% (b)
		\subfigure[]{\label{fig:d2}
			\begin{tikzpicture}[scale=\figscale, font=\figfontsize] 
				% grid
				\draw [thick, draw=black, fill=white] (0,0) grid (4,10) rectangle (0,0);
				\draw [thick, draw=black, fill=black] (0,0) rectangle (1,5);
				\draw [thick, draw=black, fill=black] (1,3) rectangle (2,8);
				% x axis label
				\foreach \i in {1,...,4} {
					\draw (\i,1) -- (\i,10) node [below] at (\i-0.5,0) {$\i$};
				}
				\node[] at (2,-1.5) {gene no.};
			\end{tikzpicture}
		}%
		% (c)
		\subfigure[]{\label{fig:d3}
			\begin{tikzpicture}[scale=\figscale, font=\figfontsize] 
				% grid
				\draw [thick, draw=black, fill=white] (0,0) grid (4,10) rectangle (0,0);
				\draw [thick, draw=black, fill=black] (0,0) rectangle (1,5);
				\draw [thick, draw=black, fill=black] (1,3) rectangle (2,8);
				\draw [thick, draw=black, fill=black] (2,2) rectangle (3,5);
				% x axis label
				\foreach \i in {1,...,4} {
					\draw (\i,1) -- (\i,10) node [below] at (\i-0.5,0) {$\i$};
				}
				\node[] at (2,-1.5) {gene no.};
			\end{tikzpicture}
		}%
		% (d)
		\subfigure[]{\label{fig:d4}
			\begin{tikzpicture}[scale=\figscale, font=\figfontsize] 
				% grid
				\draw [thick, draw=black, fill=white] (0,0) grid (4,10) rectangle (0,0);
				\draw [thick, draw=black, fill=black] (0,0) rectangle (1,5);
				\draw [thick, draw=black, fill=black] (1,3) rectangle (2,8);
				\draw [thick, draw=black, fill=black] (2,2) rectangle (3,5);
				\draw [thick, draw=black, fill=black] (3,2) rectangle (4,4);
				% x axis label
				\foreach \i in {1,...,4} {
					\draw (\i,1) -- (\i,10) node [below] at (\i-0.5,0) {$\i$};
				}
				\node[] at (2,-1.5) {gene no.};
			\end{tikzpicture}
		}
	\caption{Translation of $x$-$y$-axis genome [5,8,2,4]. In (a) the voxels are enabled from the first allele (5) to the baseline (bottom). Subsequently in (b) the voxels are enabled from the second allele (8) to the previous allele (5) and extended 2 voxels. In (c) the third allele (2) is less than the previous lower position (3), causing a gap, and is thus drawn from the allele (2) to the previous lower position and extended 2 voxels to provide structural support. In (d) the allele (4) is less than the previous upper position (5) and 2 voxels are enabled from the allele toward the baseline. }
	\label{fig:drawing}
\end{figure}
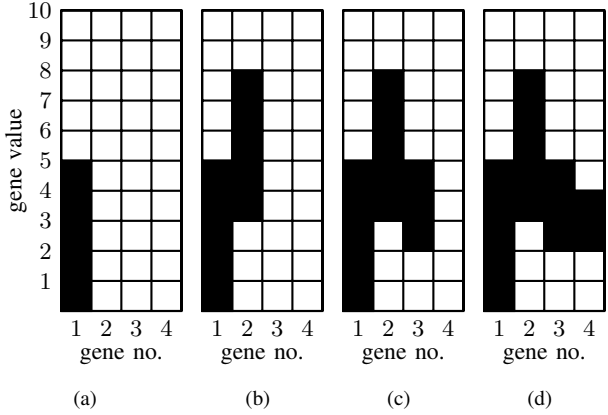

To translate the genome for a single $z$ layer, an equilateral cross is constructed using the 10 aforementioned genes, with 4 blades bent at right angles and an allele range [1,42]. For north-east and south-west quadrants the baseline is a horizontal line at $y$-axis=50, and for north-west and south-east quadrants the baseline is a vertical line at $x$-axis=50. Starting from the central platform and translating each gene successively, the one-tenth of voxels controlled by that gene are then drawn from the allele+baseline towards the baseline; see Fig.~\ref{fig:d1}. If the current allele+baseline is greater than or the same as the previous allele+baseline, the voxels are enabled from the current allele+baseline to the previous allele+baseline and extended a further 2 voxels towards the baseline for structural support; see Fig.~\ref{fig:d2}. If the current allele+baseline is less than or the same as the previous lower ending position, causing a gap, the voxels are enabled from the current allele+baseline upwards to the previous lower position and extended a further 2 voxels; see Fig.~\ref{fig:d3}. In all other cases, 2 voxels are enabled from the current allele+baseline position towards the baseline; see Fig.~\ref{fig:d4}. 

\begin{figure}[t]
	\centering 
	\subfigure{\includegraphics[width=\figwidth]{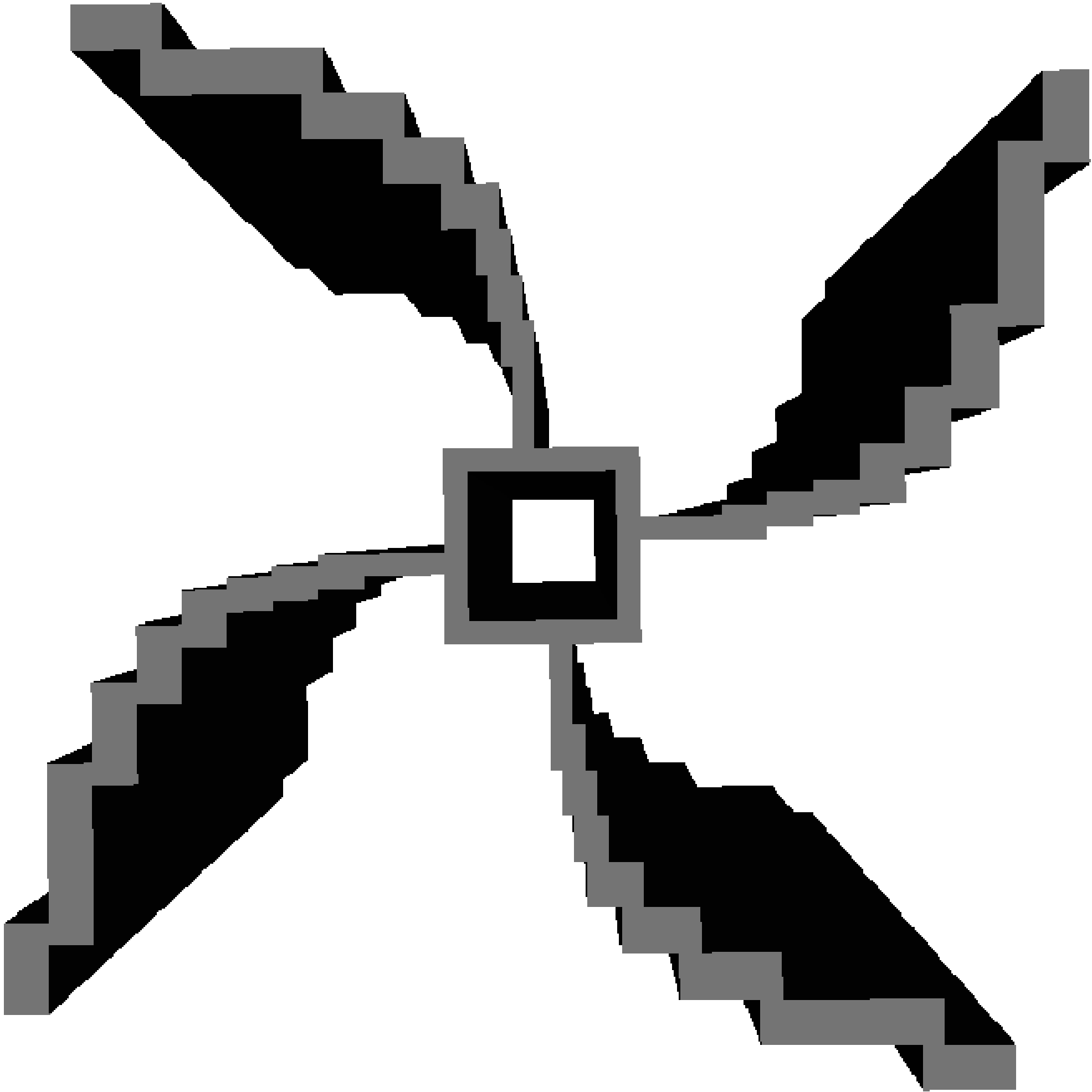} } \hspace{0.2in}
	\subfigure{\includegraphics[width=\figwidth]{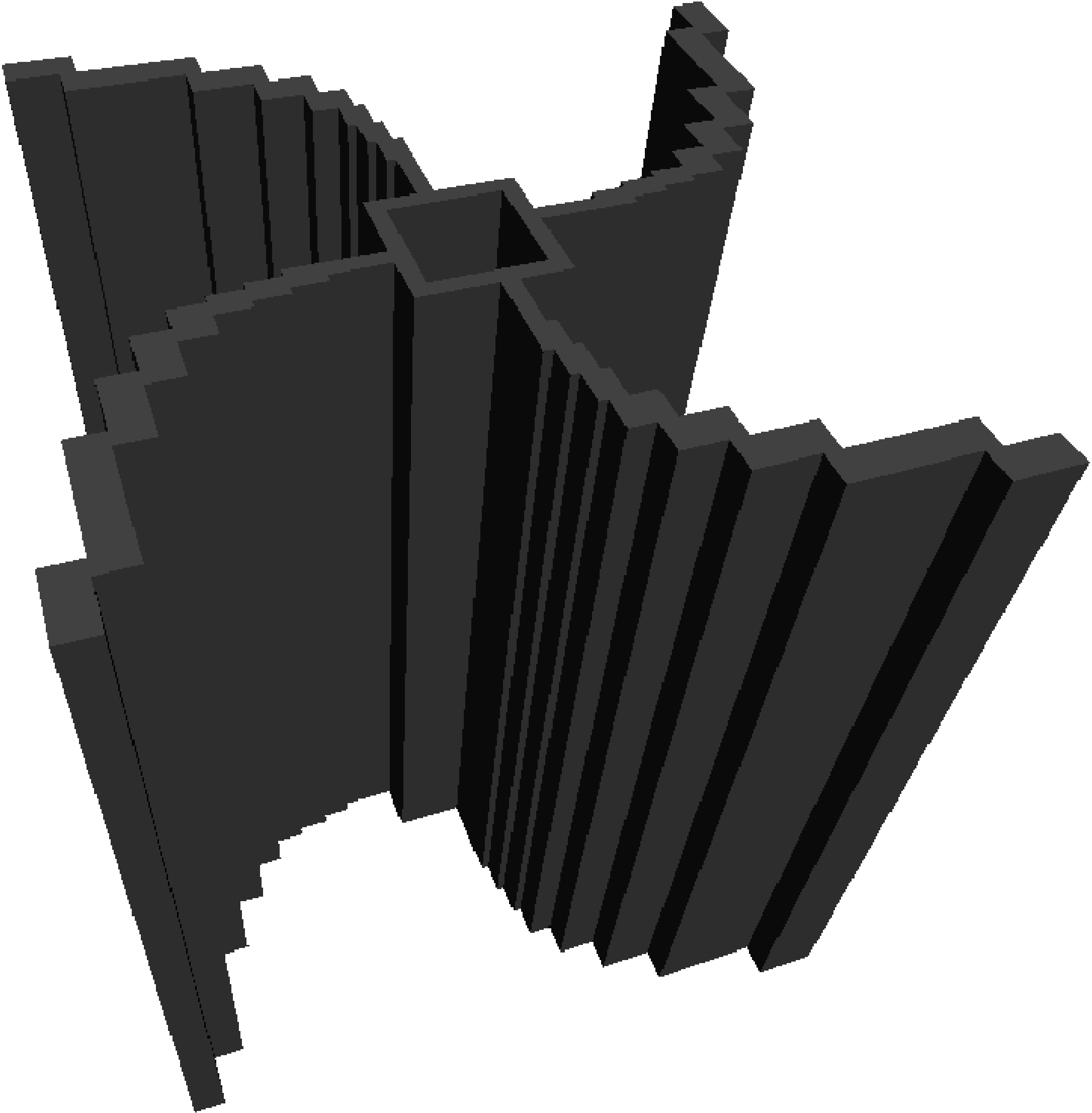} }
	\caption{Example phenotype; genome = [2, 2, 3, 4, 5, 8, 13, 20, 34, 40].}\label{fig:target-unsmoothed}
	\subfigure{\includegraphics[width=\figwidth]{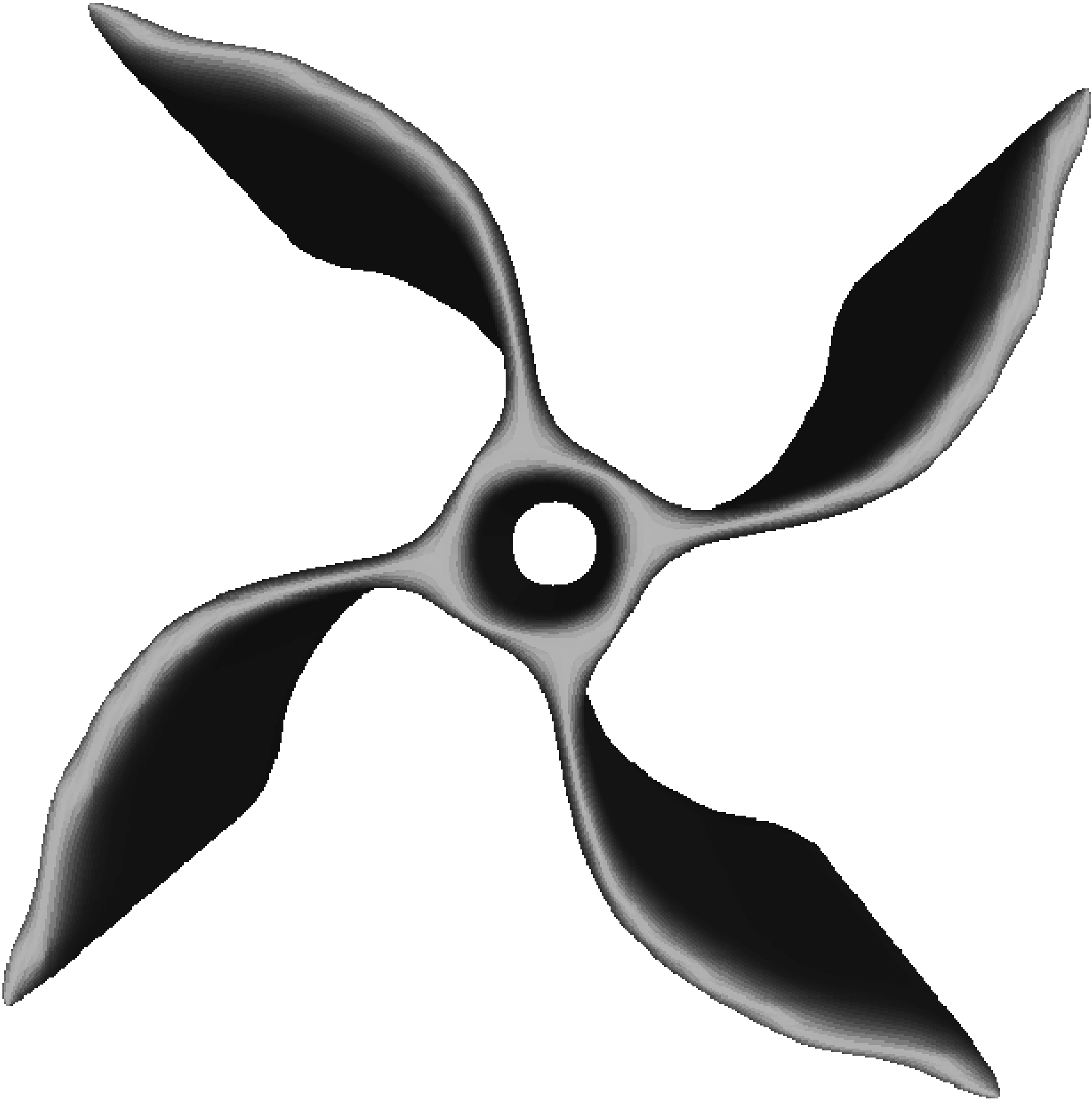} } \hspace{0.2in}
	\subfigure{\includegraphics[width=\figwidth]{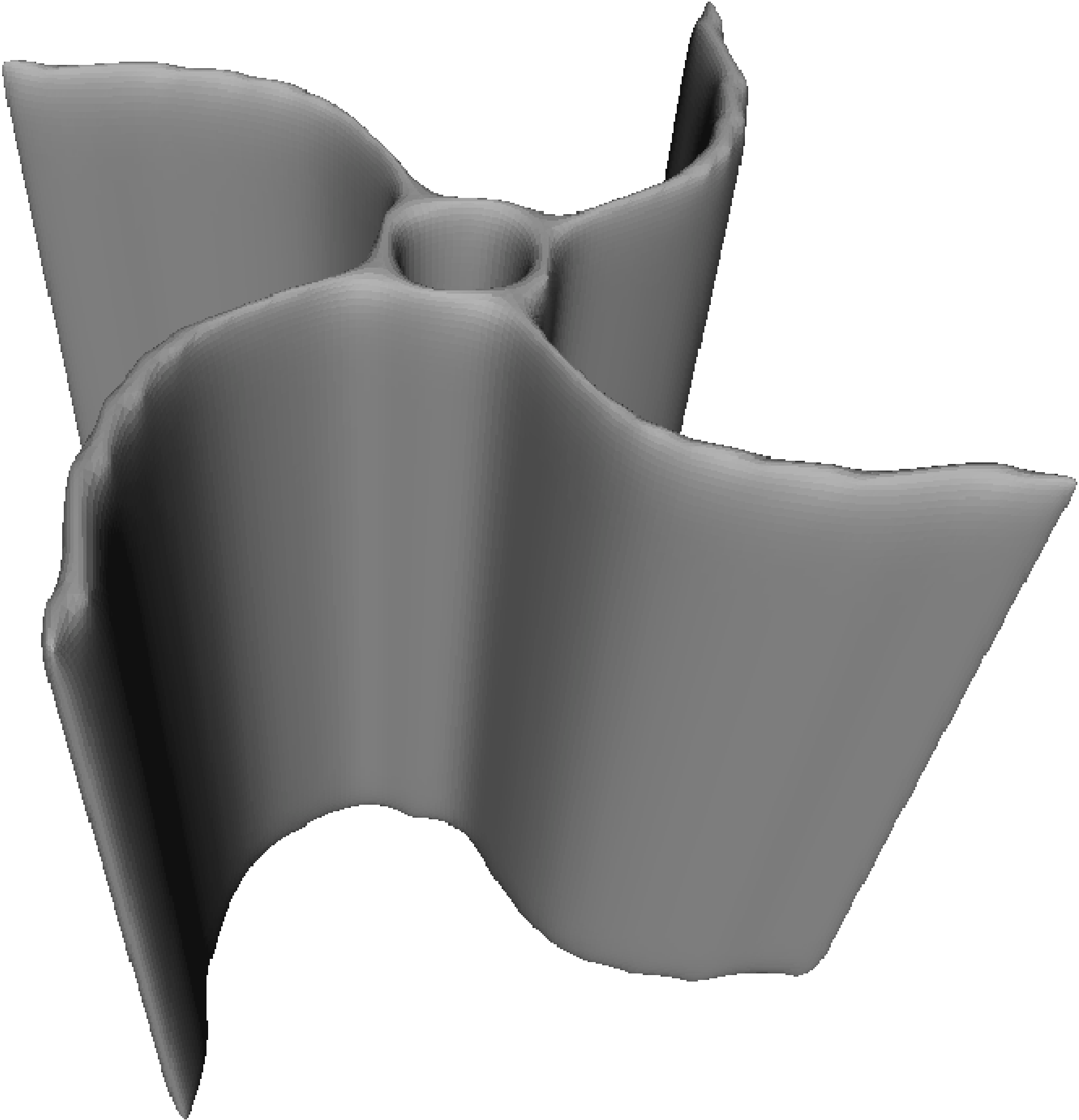} }
	\caption{Example with 50 Laplacian smoothing steps applied.}\label{fig:target-smoothed}
	\subfigure{\includegraphics[width=1.6in]{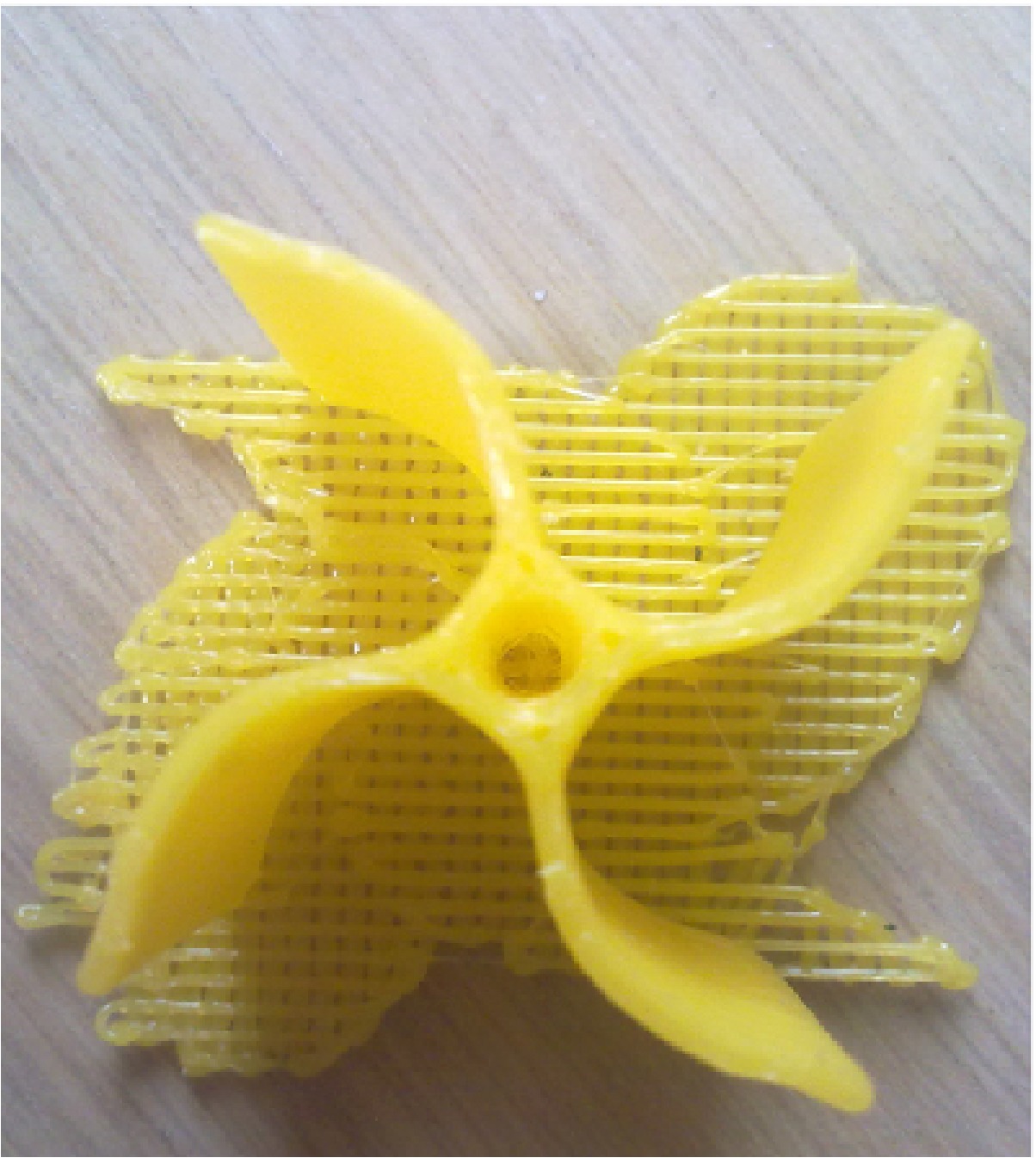} } \hspace{-0.1in}
	\subfigure{\includegraphics[width=1.6in]{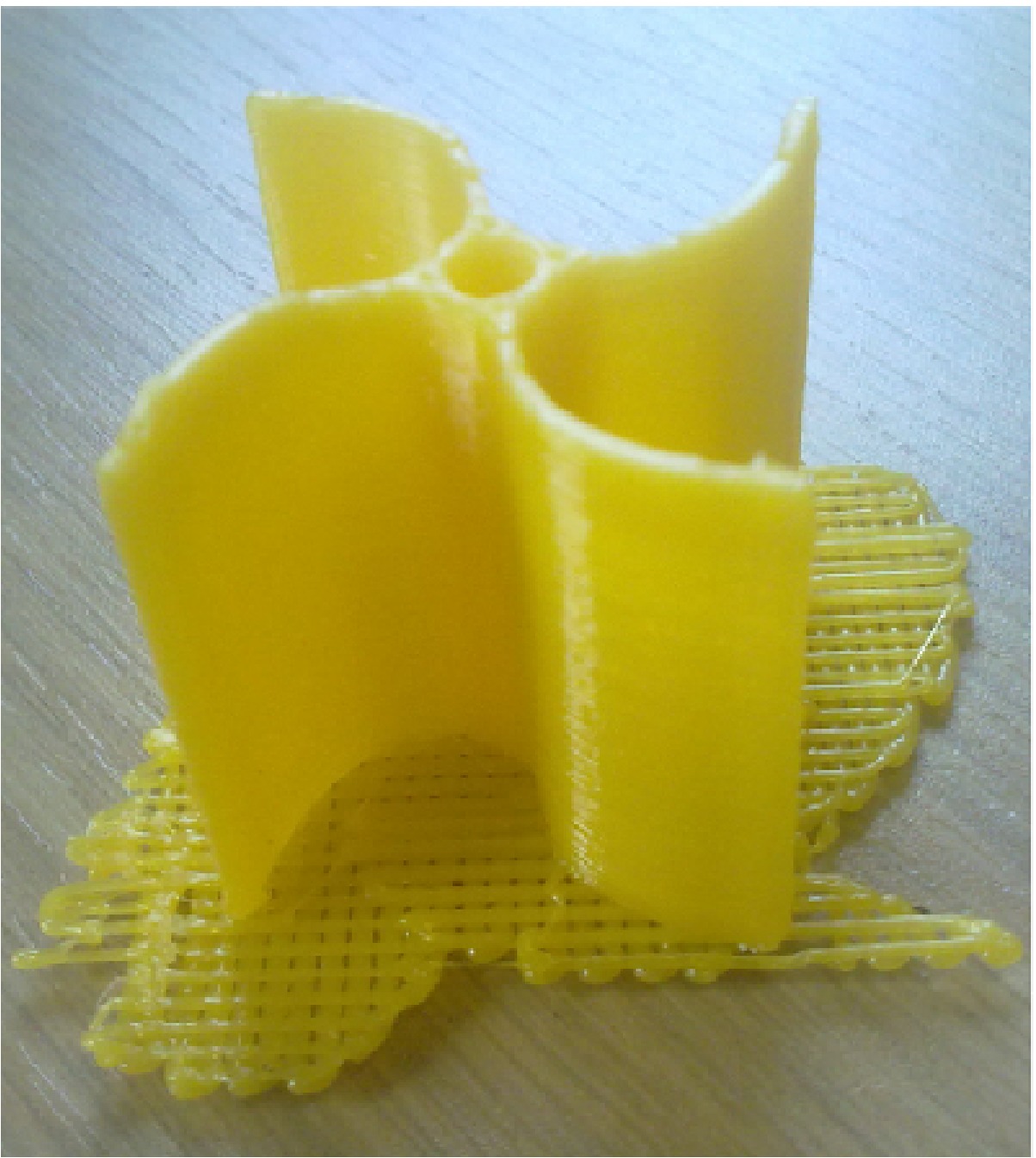} }
	\caption{Example smoothed and printed by a 3D printer; $30\times30\times30mm$; $27mins$ printing time.}\label{fig:target-printed}
\end{figure}

For non-varying $z$-axis prototypes, once the base voxel layer is constructed it is simply duplicated to fill the workspace in the $z$-dimension; an example phenotype is shown in Fig.~\ref{fig:target-unsmoothed}. When production is desired, the 3D binary voxel array is converted to stereolithography ({STL}) format. Once encoded in {STL}, it then undergoes post-processing with the application of 50 Laplacian smoothing steps using Meshlab\footnote{MeshLab is an open source, portable, and extensible system for the processing and editing of unstructured 3D triangular meshes.\ http://meshlab.sourceforge.net}; see smoothed example phenotype in Fig.~\ref{fig:target-smoothed}. Finally the object is converted to printer-readable {G}-code and is subsequently fabricated on a {BFB} 3000 printer using a polylactic acid ({PLA}) bioplastic. Fig.~\ref{fig:target-printed} shows the smoothed object after fabrication.

To enable prototypes with $z$-axis variability, the genome is extended to include 5 additional integers in the range [-42,42], each controlling 1/6th of the $z$-axis. After drawing the initial $z$-layer as previously described, each $z$ gene transforms the $x$-$y$ genome for the next successive $z$-layer by uniformly adding the allele value, after which it is then drawn as described above. For example, with an $x$-$y$-axis genome of [2, 2, 3, 4, 5, 8, 13, 20, 34, 40] and $z$-axis genome of [2, -5, 10, 3, -2], the next $z$-layer is translated using the $x$-$y$-axis genome of [4, 4, 5, 6, 7, 10, 15, 22, 36, 42] and the following $z$-layer is translated with [1, 1, 1, 1, 2, 5, 10, 17, 31, 37], etc. The surrogate model is necessarily extended from 10 input neurons to 15 to incorporate the additional genes.

The genetic algorithm (GA)~\cite{Holland:1975} used herein proceeds with a population of 20 randomly seeded individuals, a maximum mutation step size of $\pm10$, a per allele mutation rate of 25\%, and a crossover rate of 0\%. A tournament size of 3 takes place for both selection and replacement. Initially, all 20 of the random designs are fabricated and evaluated.  

For the surrogate-assisted architecture used in this paper~\cite{Bull:1999}, the basic GA remains unchanged except that fitness evaluations are obtained from a forward pass of the genome through a neural network when the real fitness value is unknown. Initially the entire population is fabricated and evaluated on the real fitness function and added to an evaluated set. The model is trained using backpropagation for 1,000 epochs; where an epoch consists of randomly selecting, without replacement, an individual from the evaluated set and updating the model weights. Each generation thereafter, the individual with the highest approximated fitness as suggested by the model and a randomly chosen unevaluated individual are fabricated and evaluated on the real fitness function and added to the evaluated set, whereupon the model is iteratively retrained. The model parameters, $\beta=0.3$, $\theta=0$, $elasticity=1$, $calming~rate=1$, $momentum=0$, $elasticity~rate=0$.

\begin{figure}[t]
	\centering 
	\includegraphics[width=3.0in]{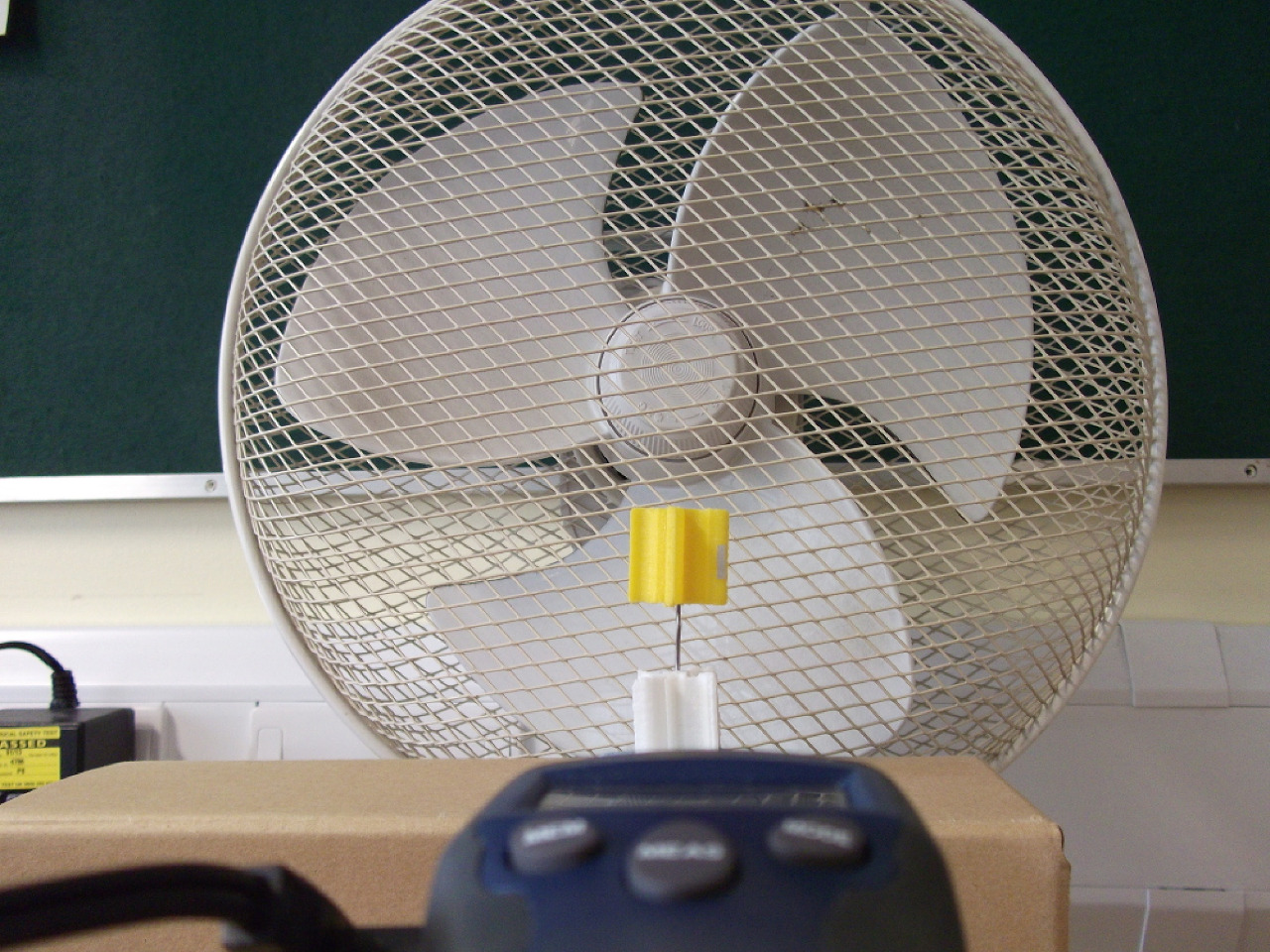}
	\caption{Single VAWT experimental setup.}
	\label{fig:setup-single}
\end{figure}

The fitness of each individual is the maximum rotation speed achieved over the period of 60 seconds during the application of constant wind generated by a propeller fan after fabrication by a 3D printer. The rotation speed is the significant measure of aerodynamic efficiency since the design space is constrained (including rotor radius and turbine height). However, in future work, the AC voltage generated will be preferred, which will take into account any slight weight variations that may affect performance. The rotation speed is here measured in number of revolutions per minute ($rpm$) using a PCE-DT62 digital photo laser tachometer by placing a $10\times2mm$ strip of reflecting tape on the outer tip of one of the individual's blades. The experimental setup for a single isolated VAWT can be seen in Fig~\ref{fig:setup-single}, which shows the $30W$, $3,500rpm$ $304.8mm$ propeller fan, which generates $4.4m/s$ wind speed, and the individual placed at $30mm$ distance from the centre, mounted on rigid metal pin $1mm$ in diameter.

\begin{figure}[t]
	\centering 
	\includegraphics[width=\graphwidth]{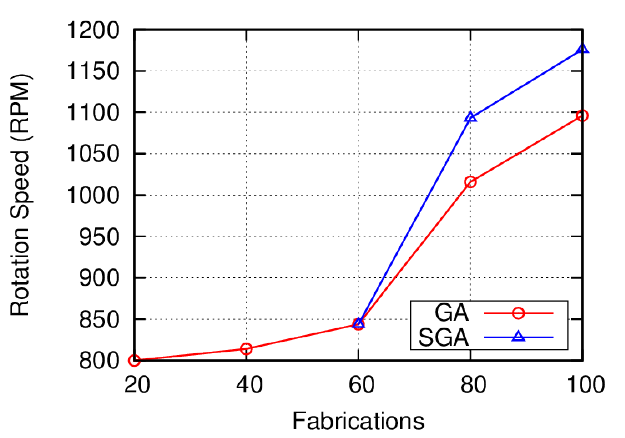}
	\caption{Rotation-based evolution. Fittest GA (circle) and SGA (triangle) individuals. The SGA is used for comparison only after 60 evaluations (i.e., 3 generations) of the GA since the initial designs are extremely aerodynamically inefficient and sufficient training data is required for the surrogate model.}
	\label{fig:real-res}
\end{figure}

\begin{figure}[t]
	\centering 
	\subfigure[1st Gen]{\label{fig:ga-g1}\includegraphics[width=\smallfigwidth]{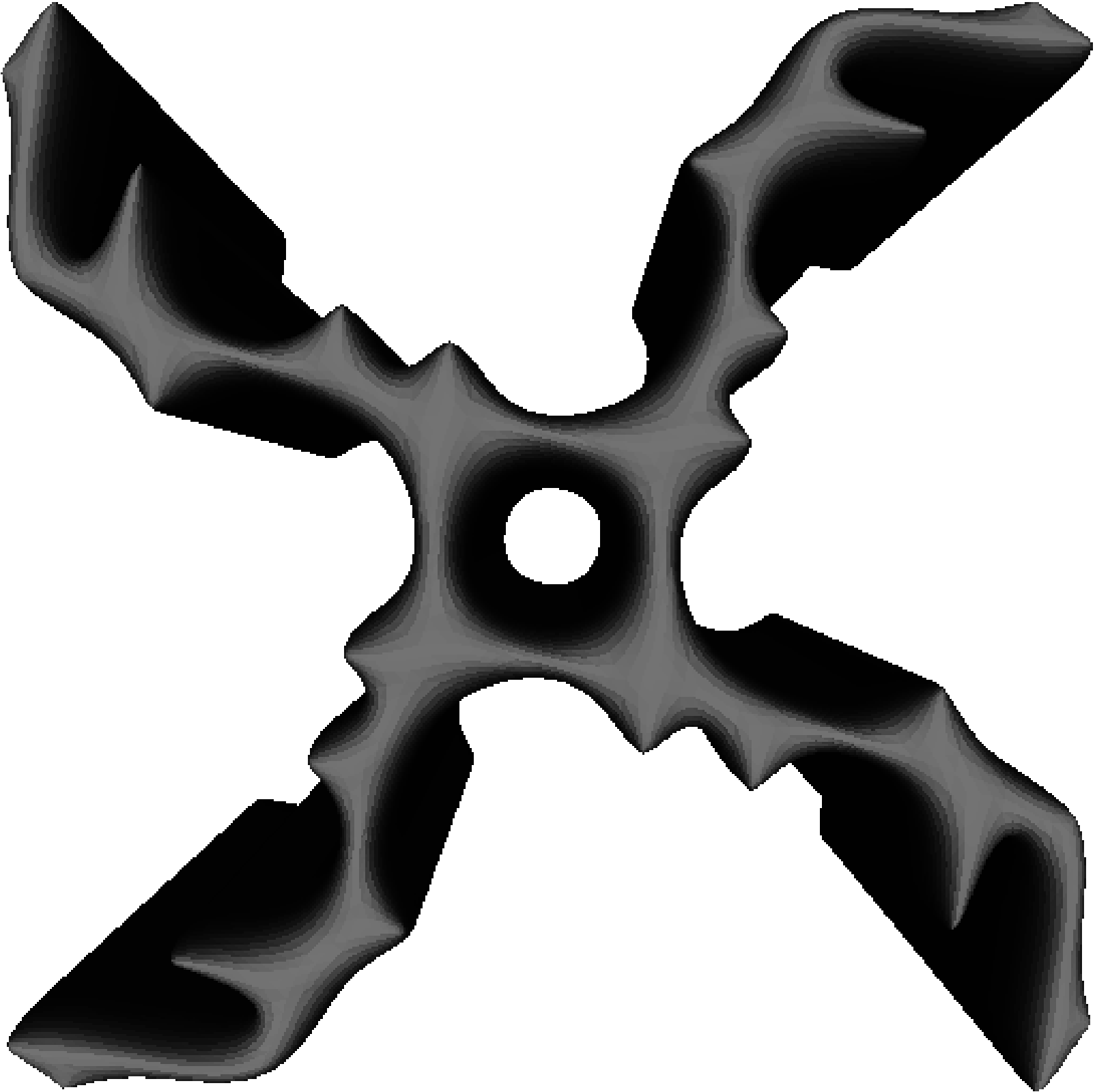}}
	\subfigure[2nd Gen]{\label{fig:ga-g2}\includegraphics[width=\smallfigwidth]{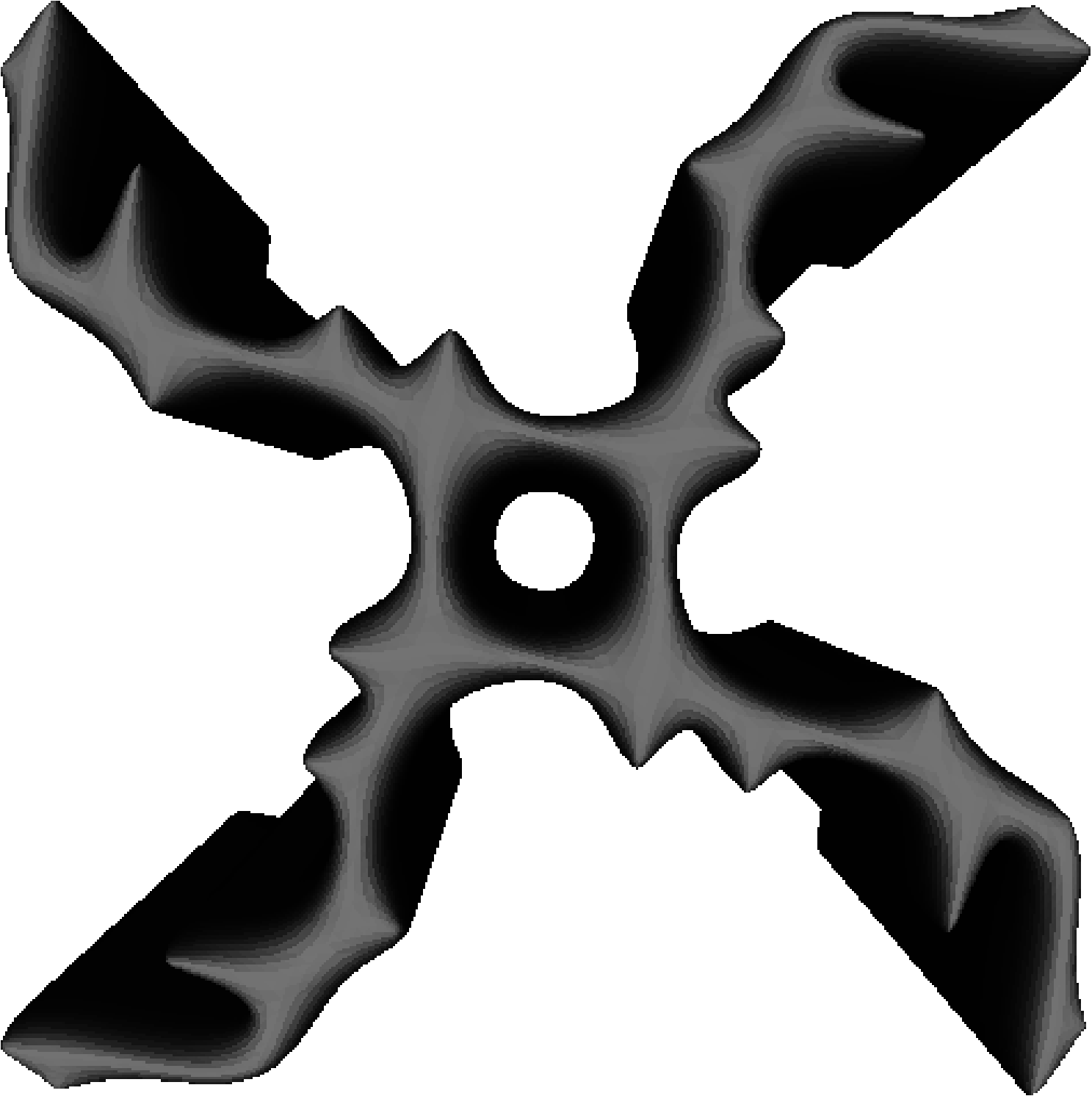}}
	\subfigure[3rd Gen]{\label{fig:ga-g3}\includegraphics[width=\smallfigwidth]{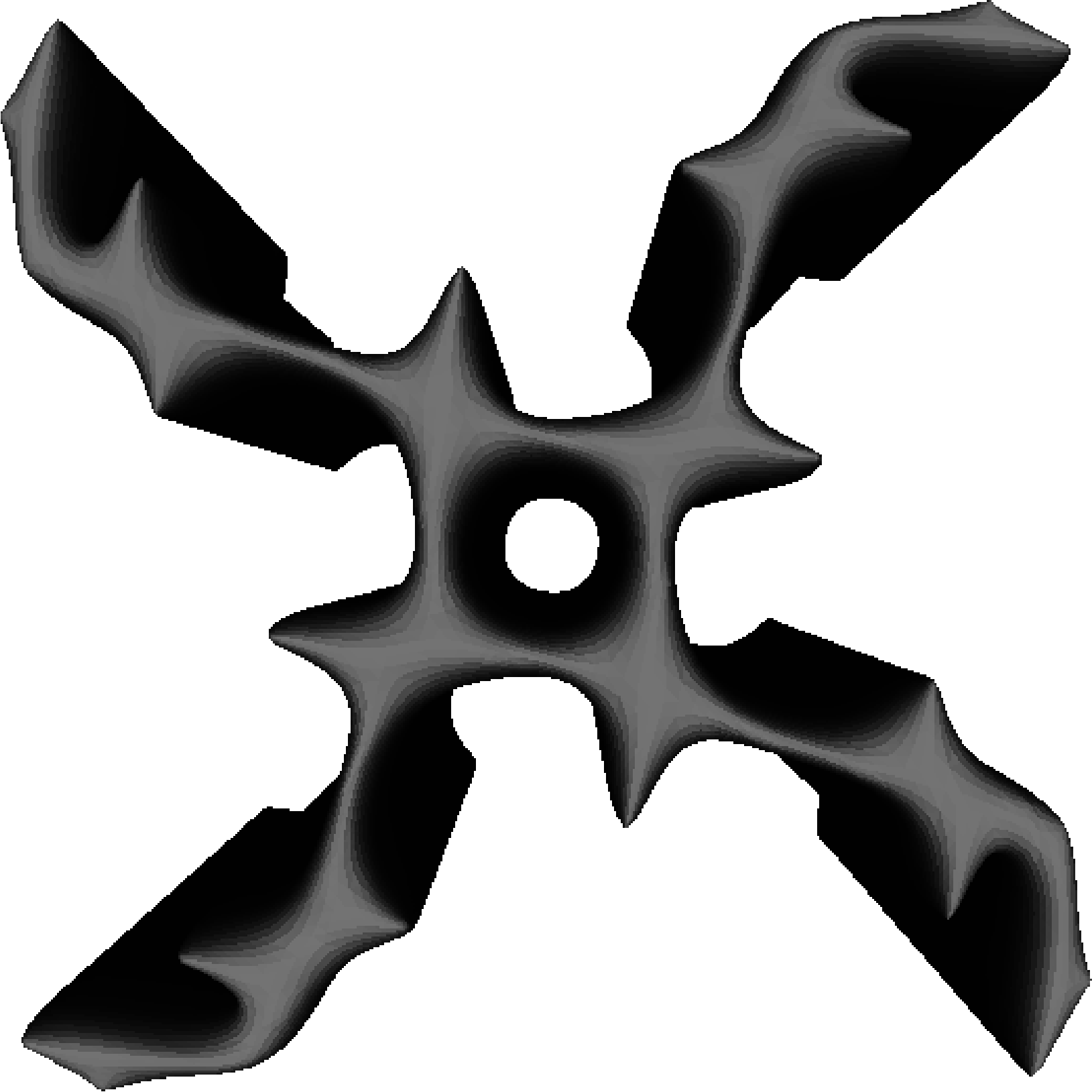}}
	\subfigure[4th Gen]{\label{fig:ga-g4}\includegraphics[width=\smallfigwidth]{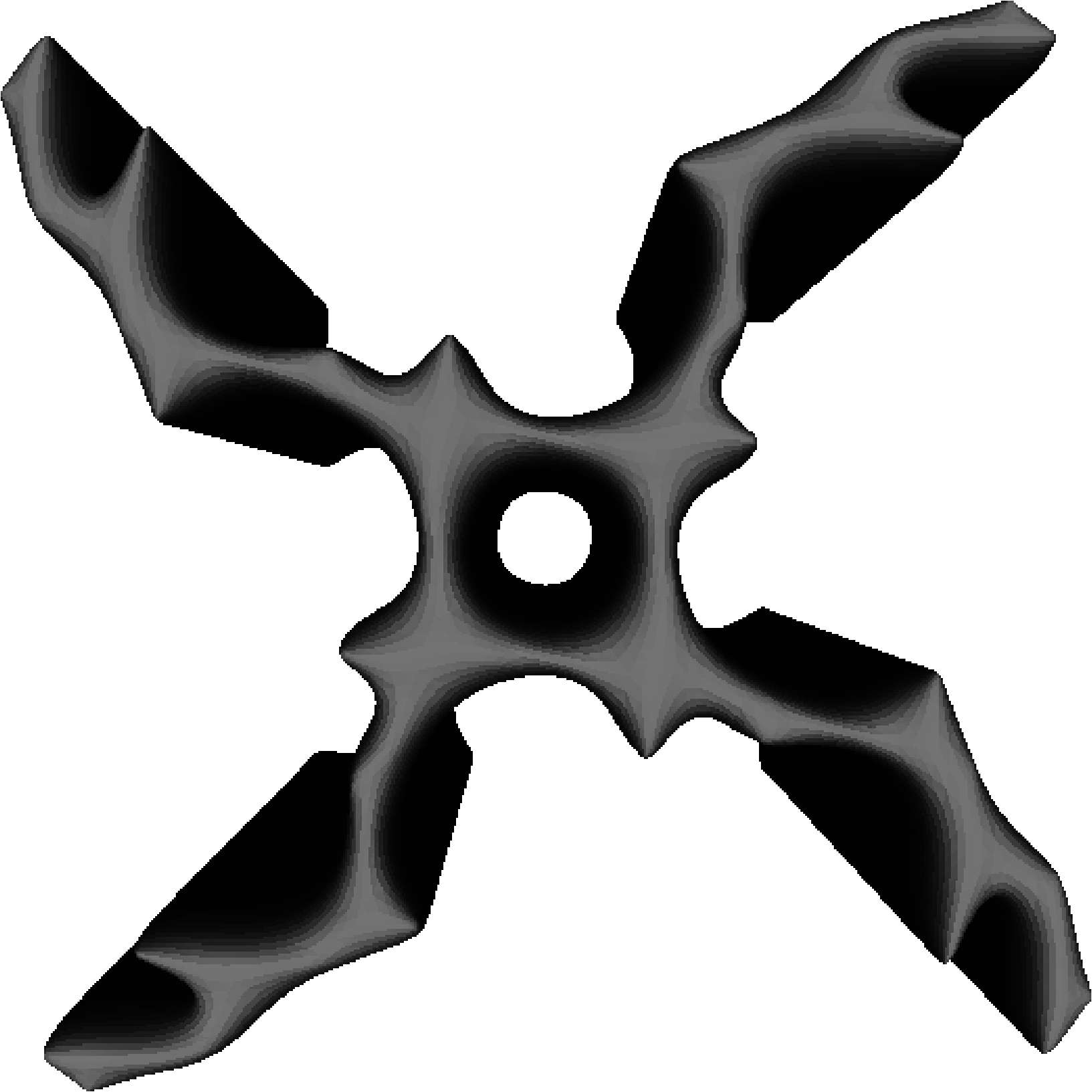}}
	\subfigure[5th Gen]{\label{fig:ga-g5}\includegraphics[width=\smallfigwidth]{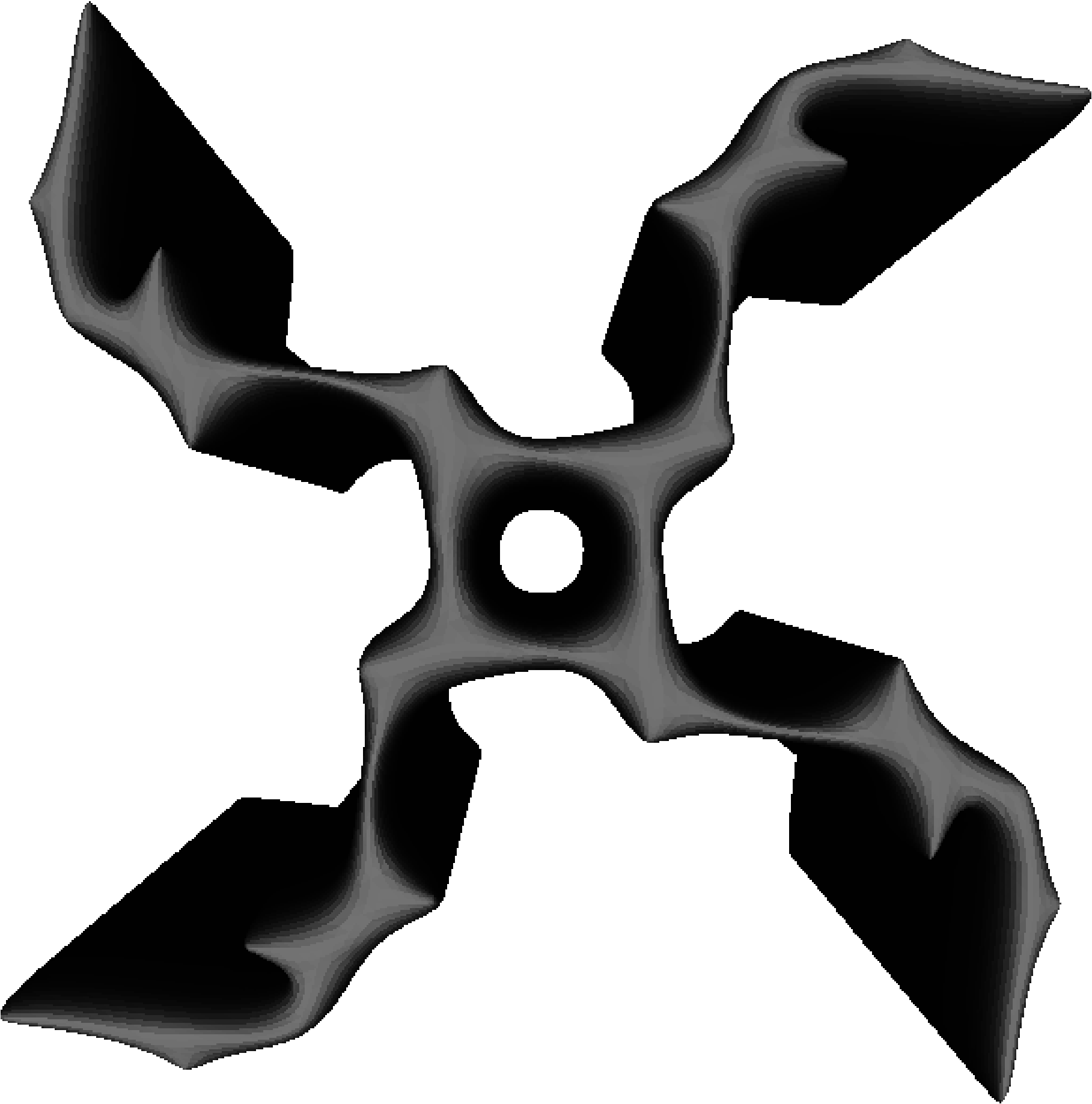}}
	\caption{The fittest evolved individuals produced by the GA each generation.}
	\label{fig:ga-individuals}
	\subfigure[4th Gen]{\label{fig:n-g4}\includegraphics[width=\smallfigwidth]{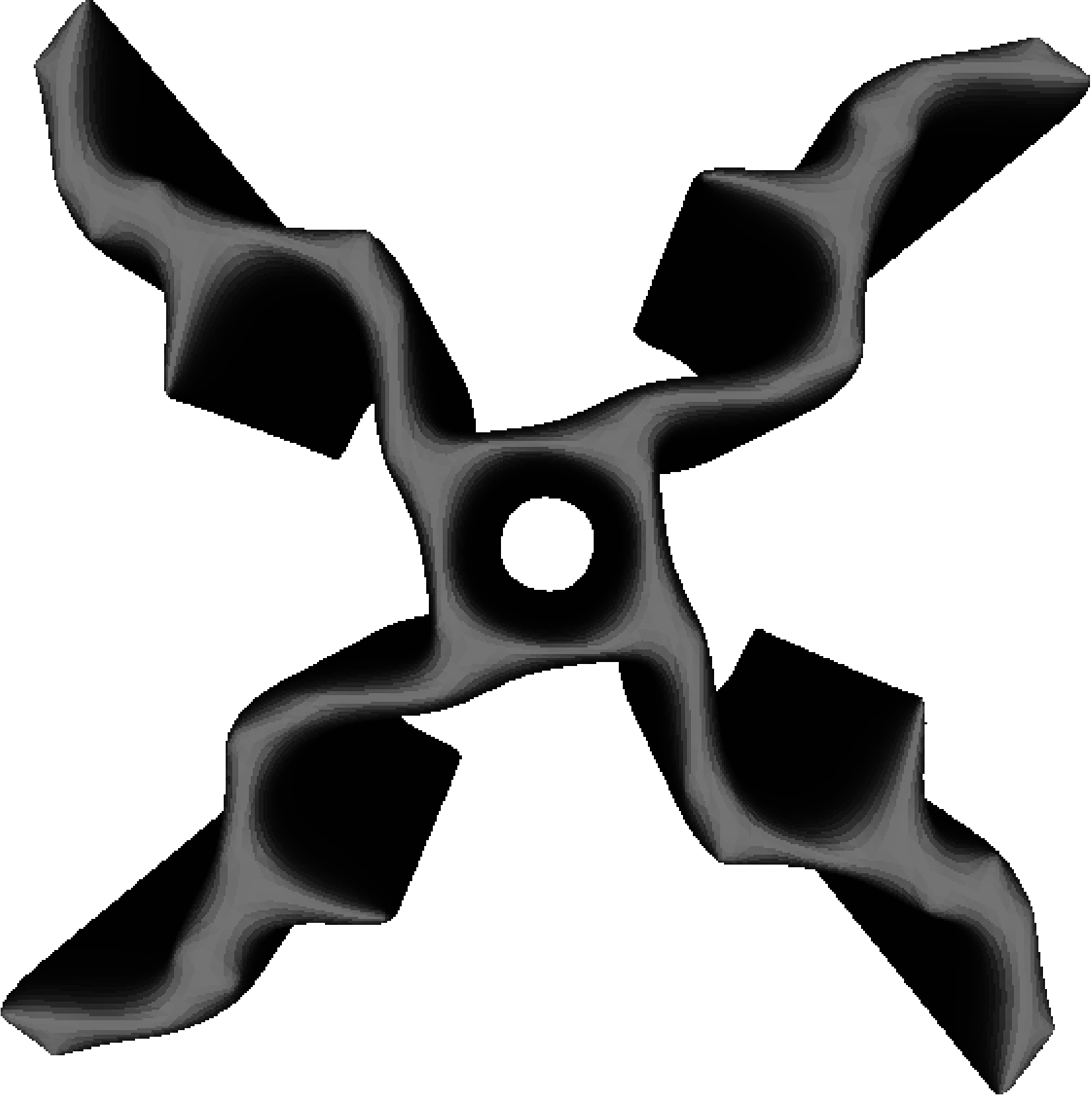}}
	\subfigure[5th Gen]{\label{fig:n-g5}\includegraphics[width=\smallfigwidth]{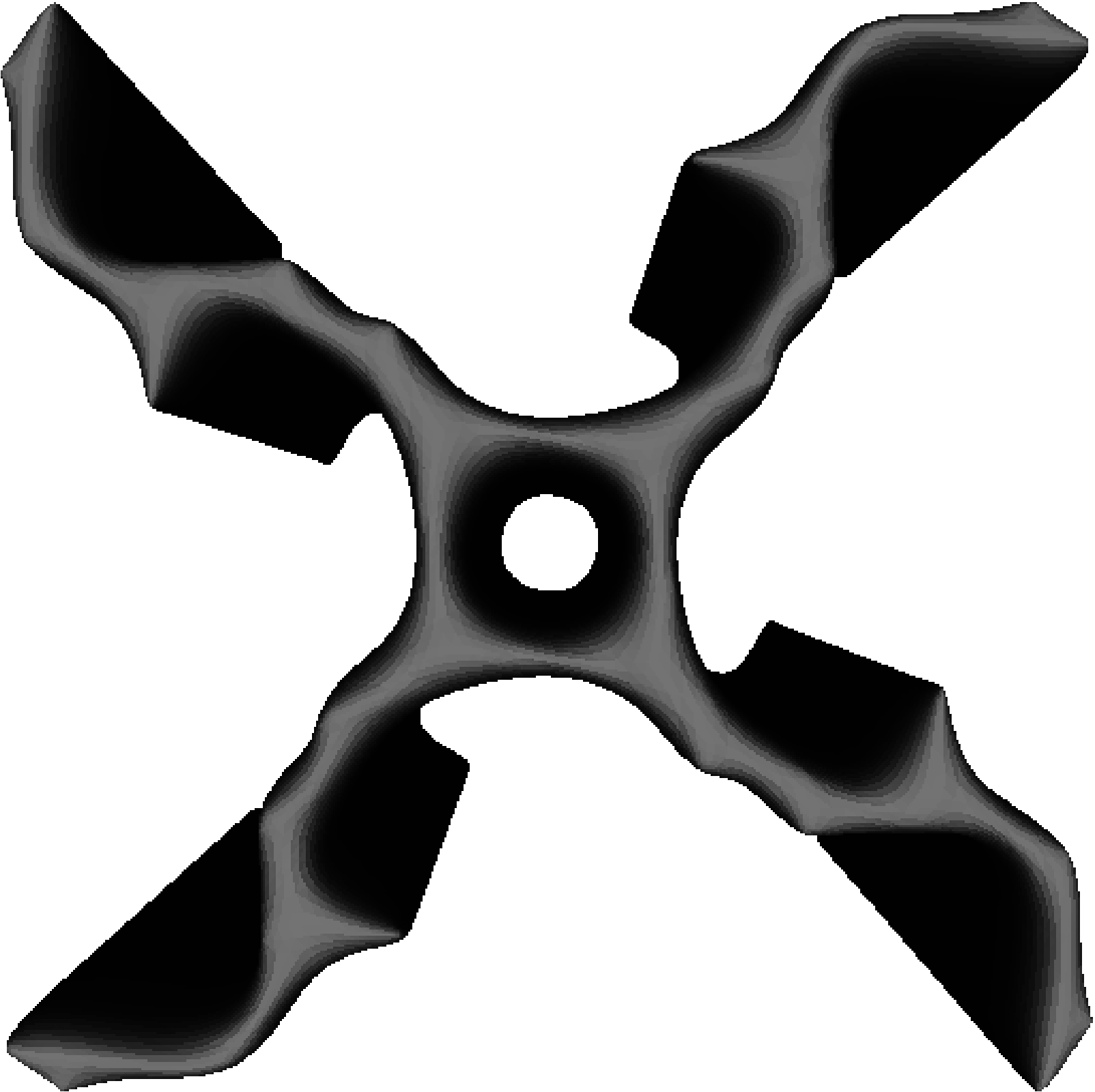}}
	\caption{The fittest evolved individuals produced by the SGA each generation.}
	\label{fig:model-individuals}
\end{figure}

Since many of the seed individuals are extremely aerodynamically inefficient (only 2 out of 20 yielded $>0rpm$), the GA is run for 2 further generations before the model-augmented approach is used for comparison. The initial pilot results from an experiment with the canonical GA and SGA are presented in Fig.~\ref{fig:real-res}, which shows the maximum rotation speed achieved by the fittest individual in each generation. The GA and model-assisted approach identify increasingly efficient aerodynamic designs, and the SGA shows improved performance ($1176rpm$ vs. $1096rpm$ after 100 fabrications). The fittest individuals produced by the GA and SGA each generation are shown in Figs.~\ref{fig:ga-individuals} and~\ref{fig:model-individuals}, respectively.
 
\begin{figure}[t]
	\centering 
	\includegraphics[width=\graphwidth]{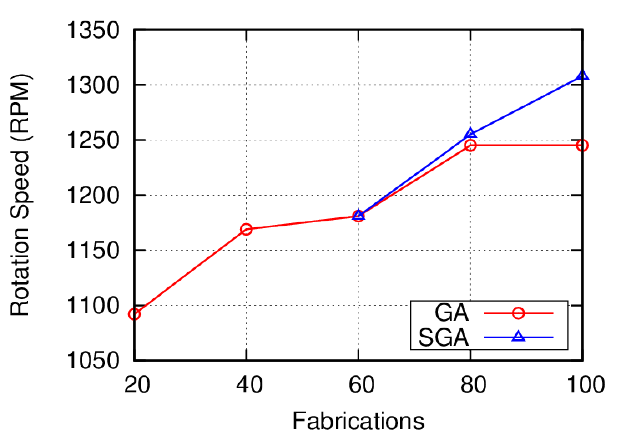}
	\caption{Rotation-based evolution with $z$-variability. Fittest GA (circle) and SGA (triangle) individuals. The SGA is used for comparison only after 60 evaluations (i.e., 3 generations) of the GA since the initial designs are extremely aerodynamically inefficient and sufficient training data is required for the surrogate model.}
	\label{fig:zreal-res}
\end{figure}

\begin{figure}[t]
	\centering 
	\subfigure[1st Gen]{\label{fig:zga-g1}\includegraphics[width=\zsmallfigwidth]{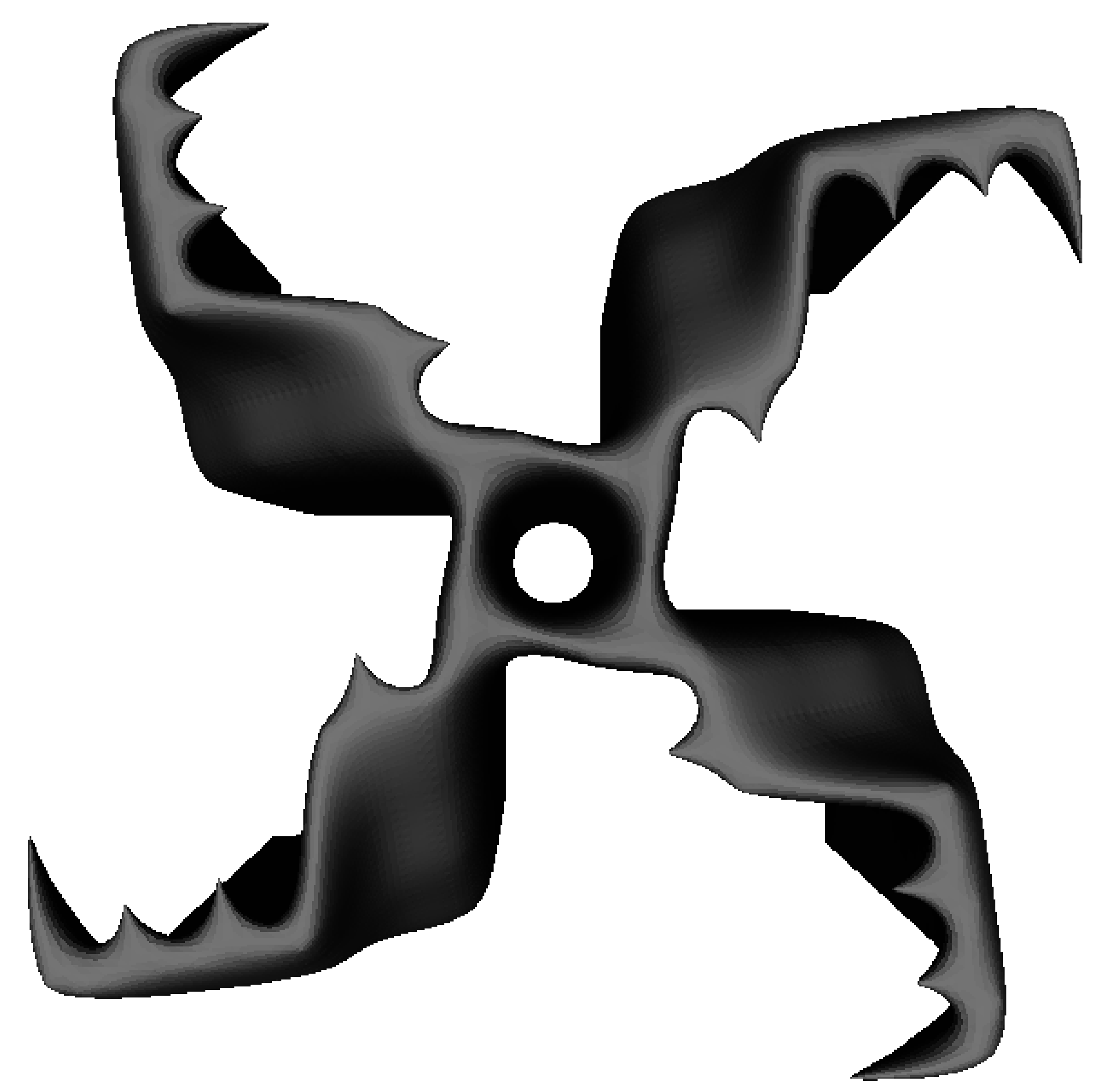} \includegraphics[width=\zsmallfigwidth]{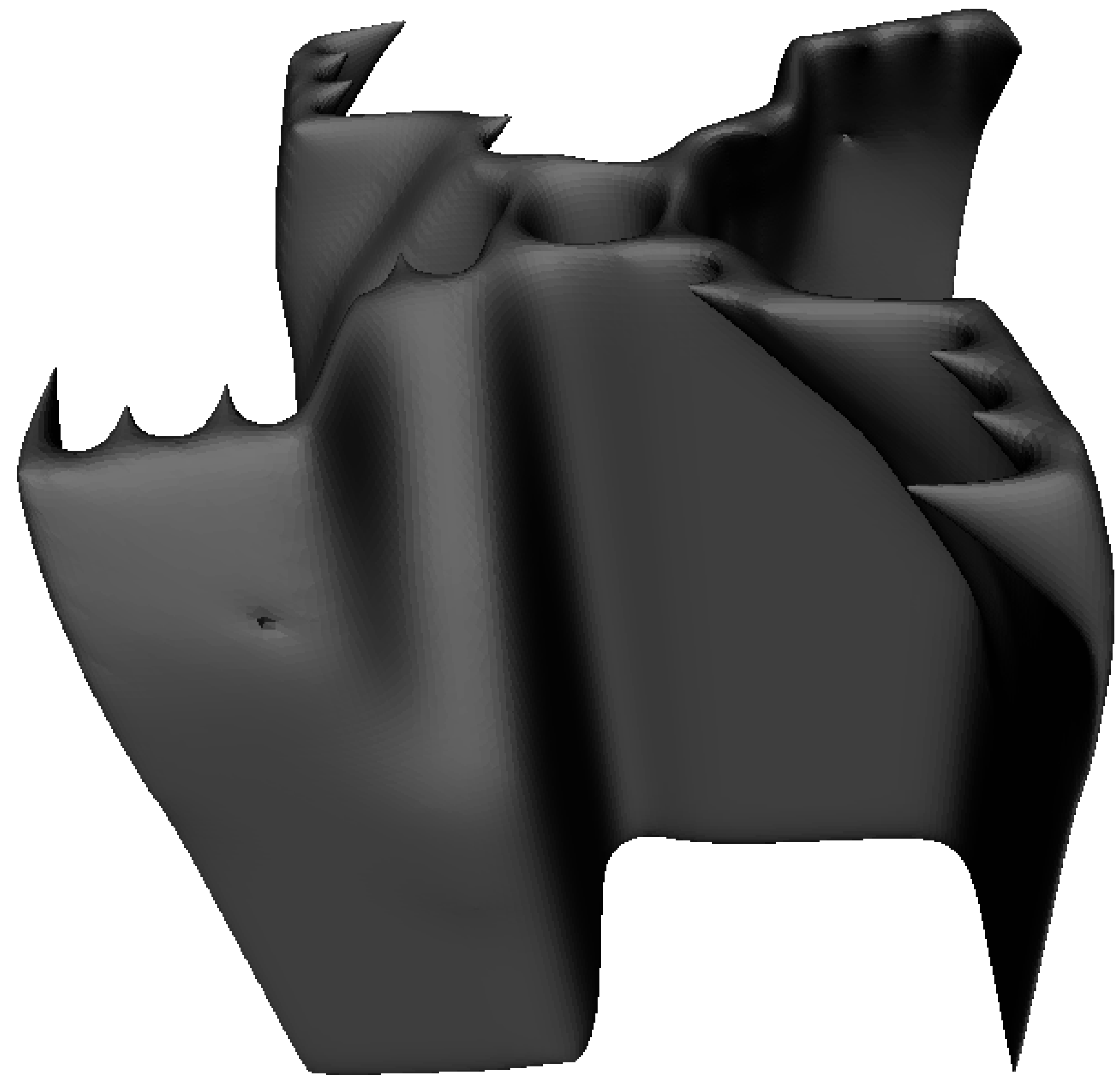}}
	\subfigure[2nd Gen]{\label{fig:zga-g2}\includegraphics[width=\zsmallfigwidth]{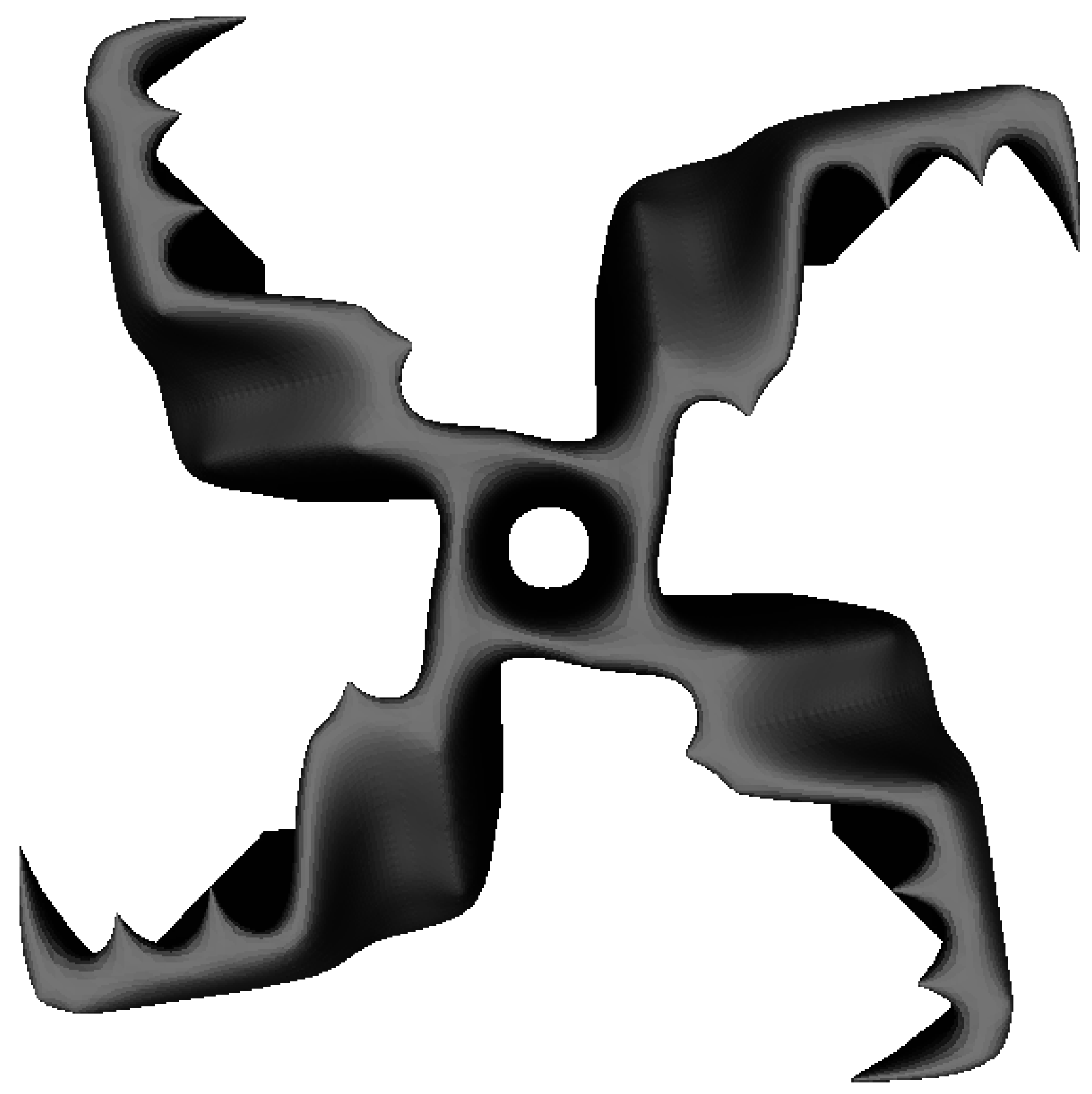} \includegraphics[width=\zsmallfigwidth]{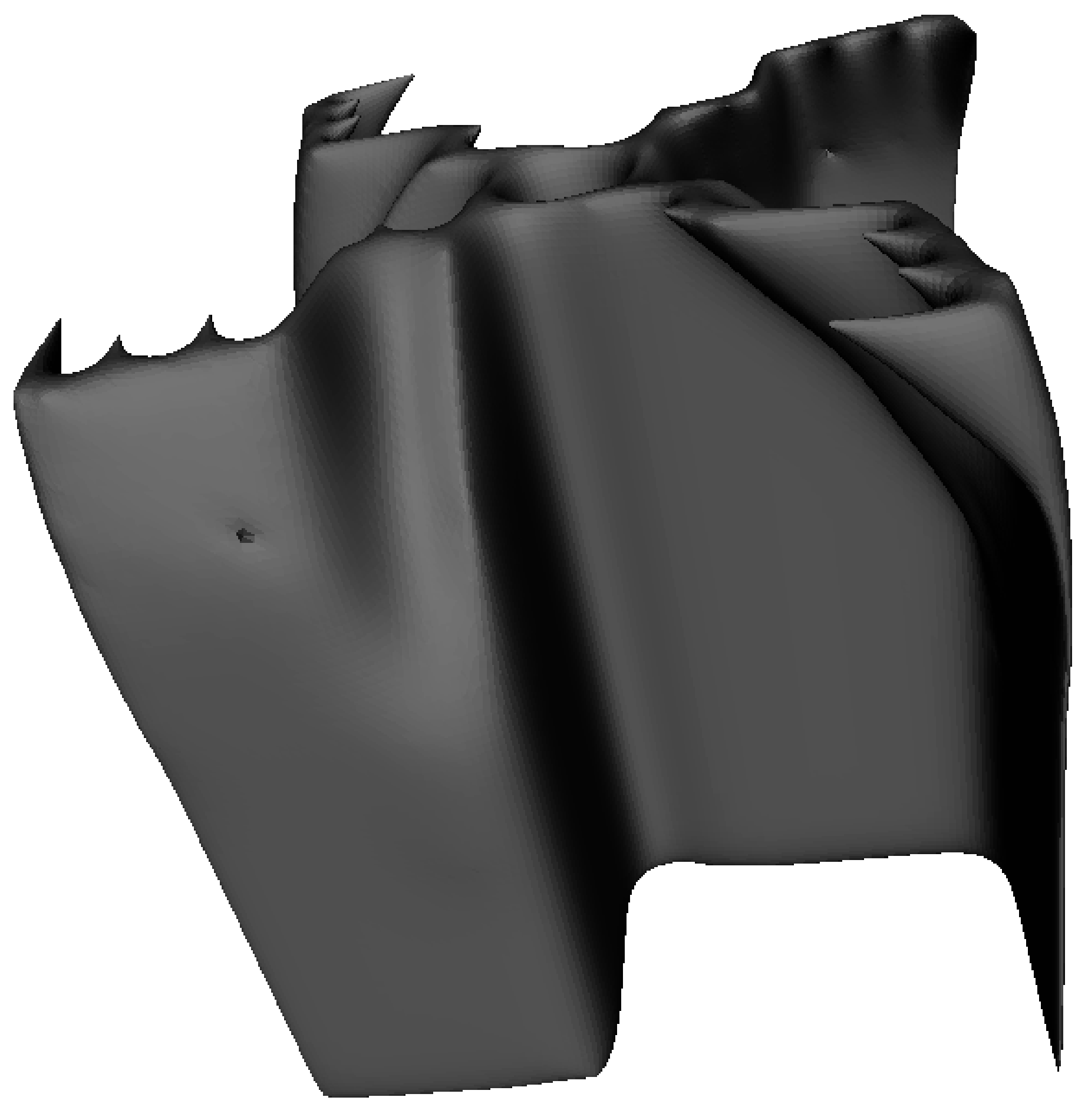}}
	\subfigure[3rd Gen]{\label{fig:zga-g3}\includegraphics[width=\zsmallfigwidth]{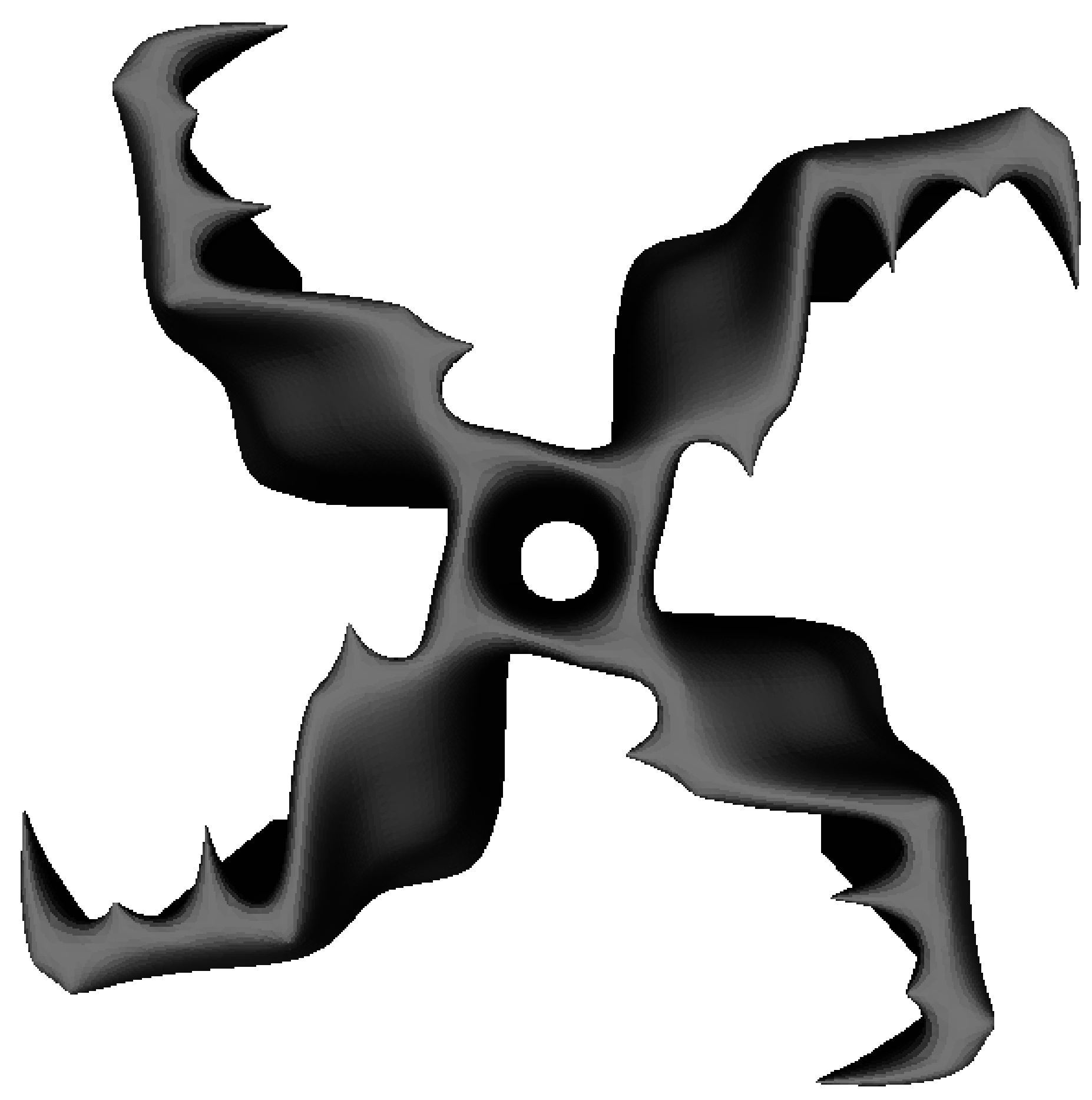} \includegraphics[width=\zsmallfigwidth]{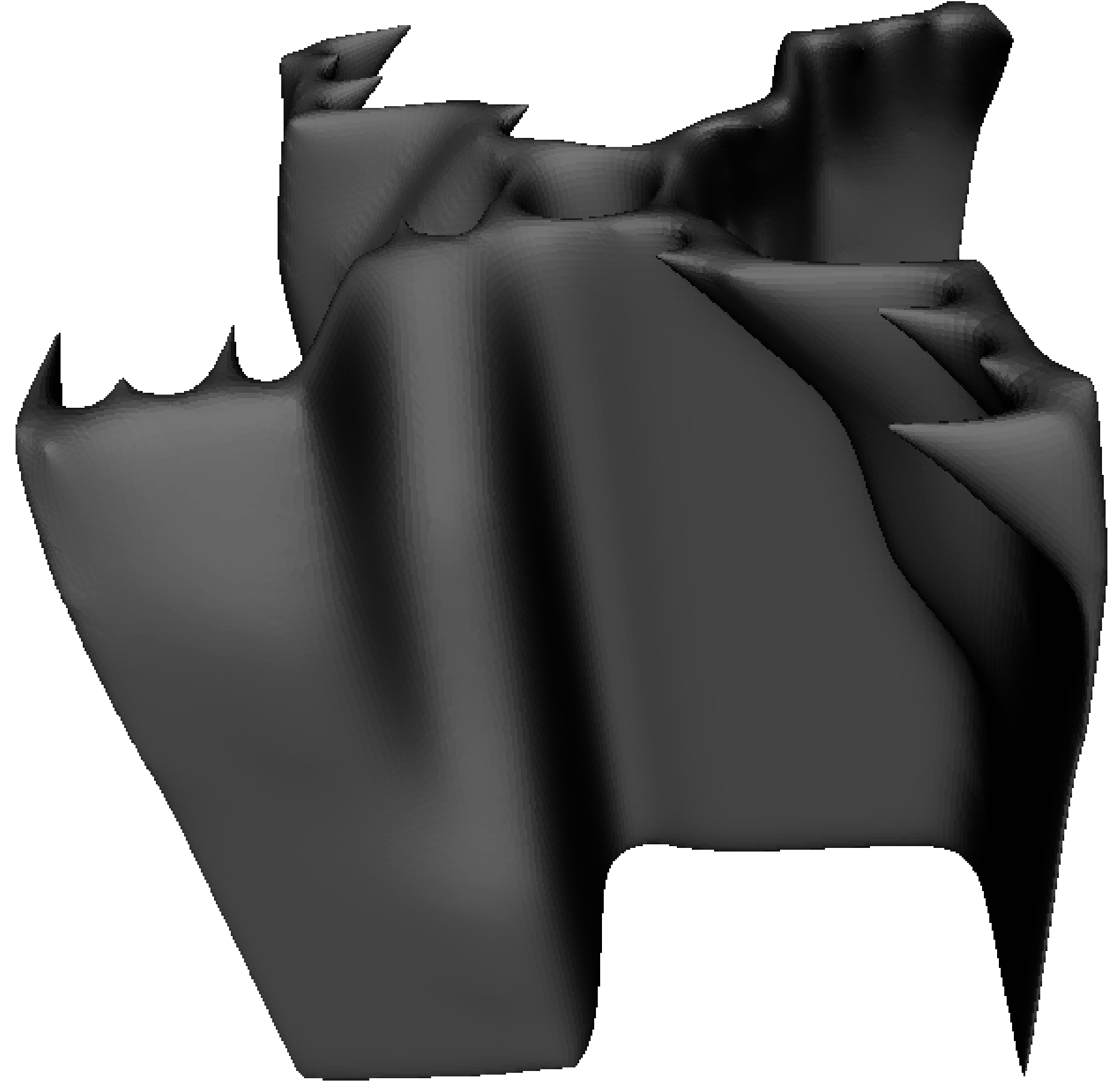}}
	\subfigure[4th/5th Gen]{\label{fig:zga-g45}\includegraphics[width=\zsmallfigwidth]{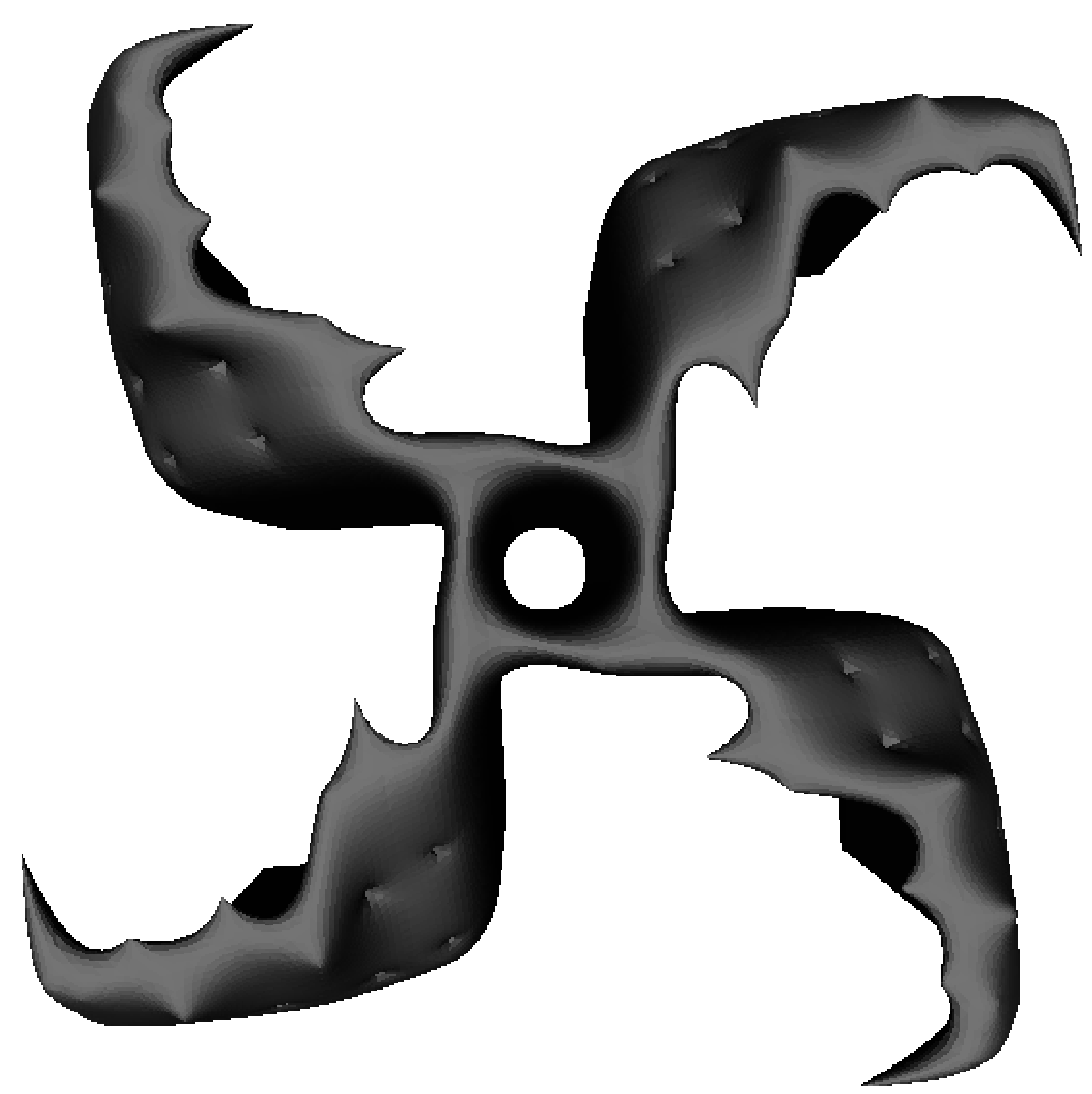} \includegraphics[width=\zsmallfigwidth]{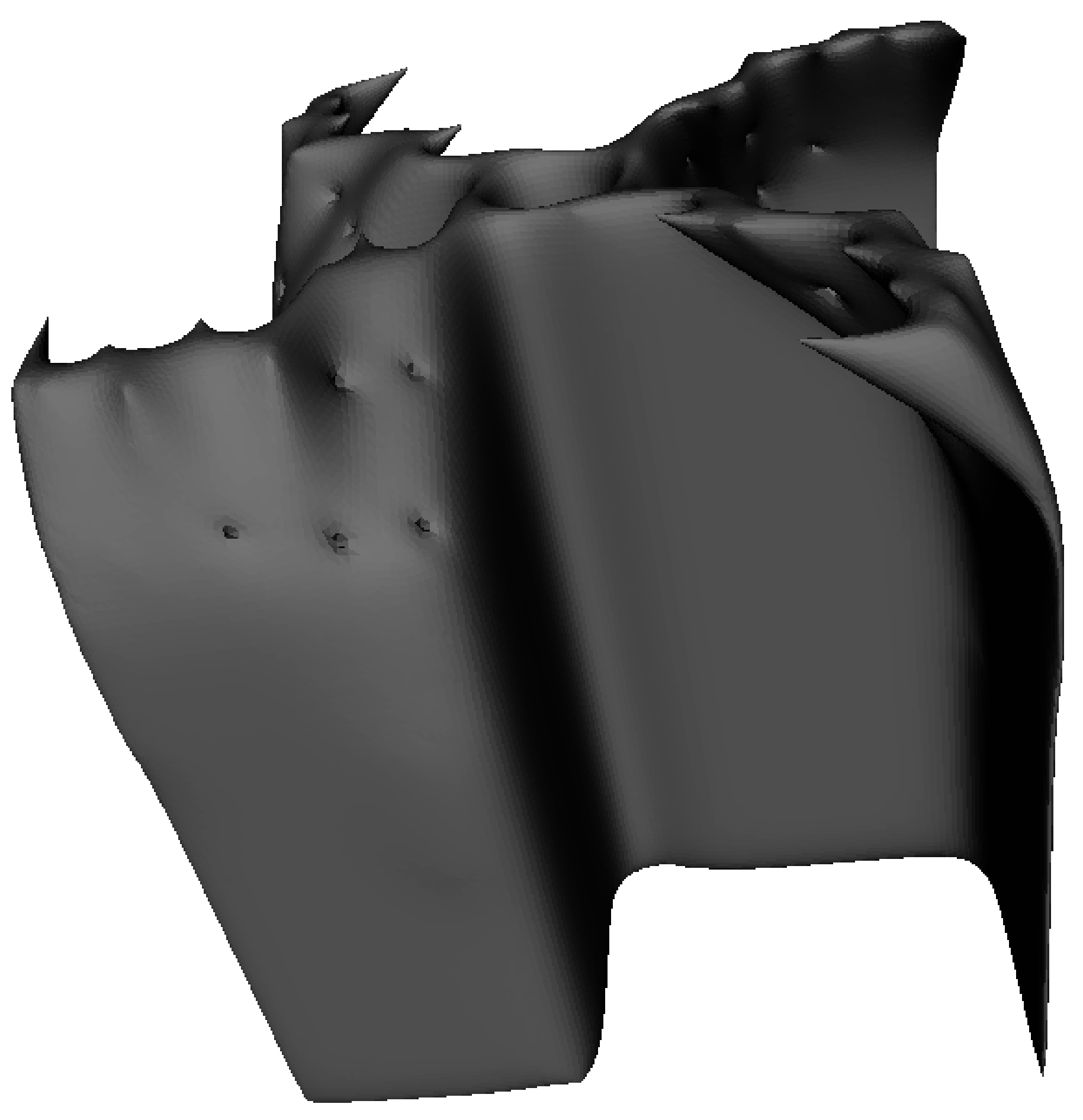}}
	\caption{The fittest evolved individuals with $z$-variability produced by the GA each generation.}
	\label{fig:zga-individuals}
\end{figure}

\begin{figure}[t]
	\centering 
	\subfigure[4th Gen]{\label{fig:zn-g4-top}\includegraphics[width=\zsmallfigwidth]{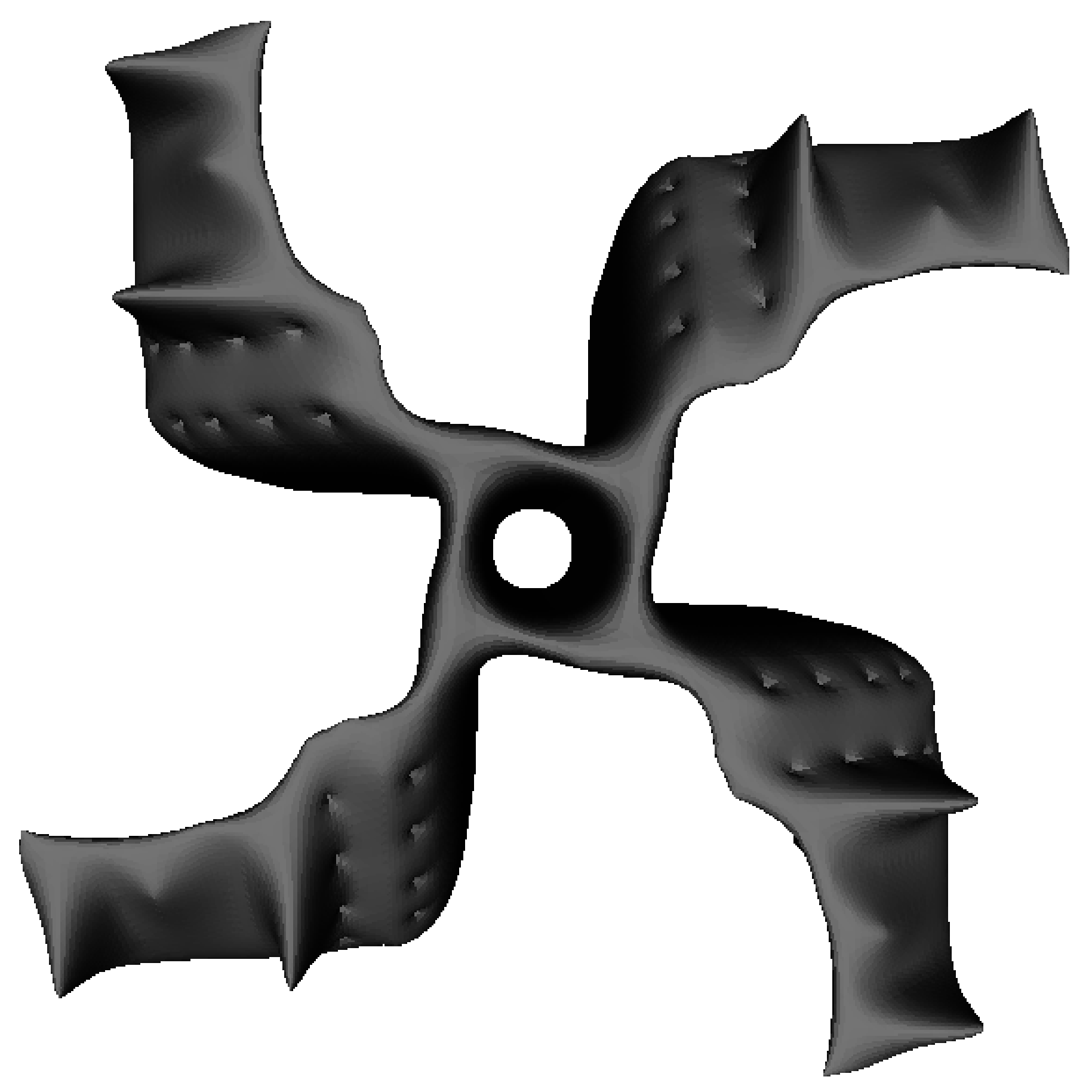} \includegraphics[width=\zsmallfigwidth]{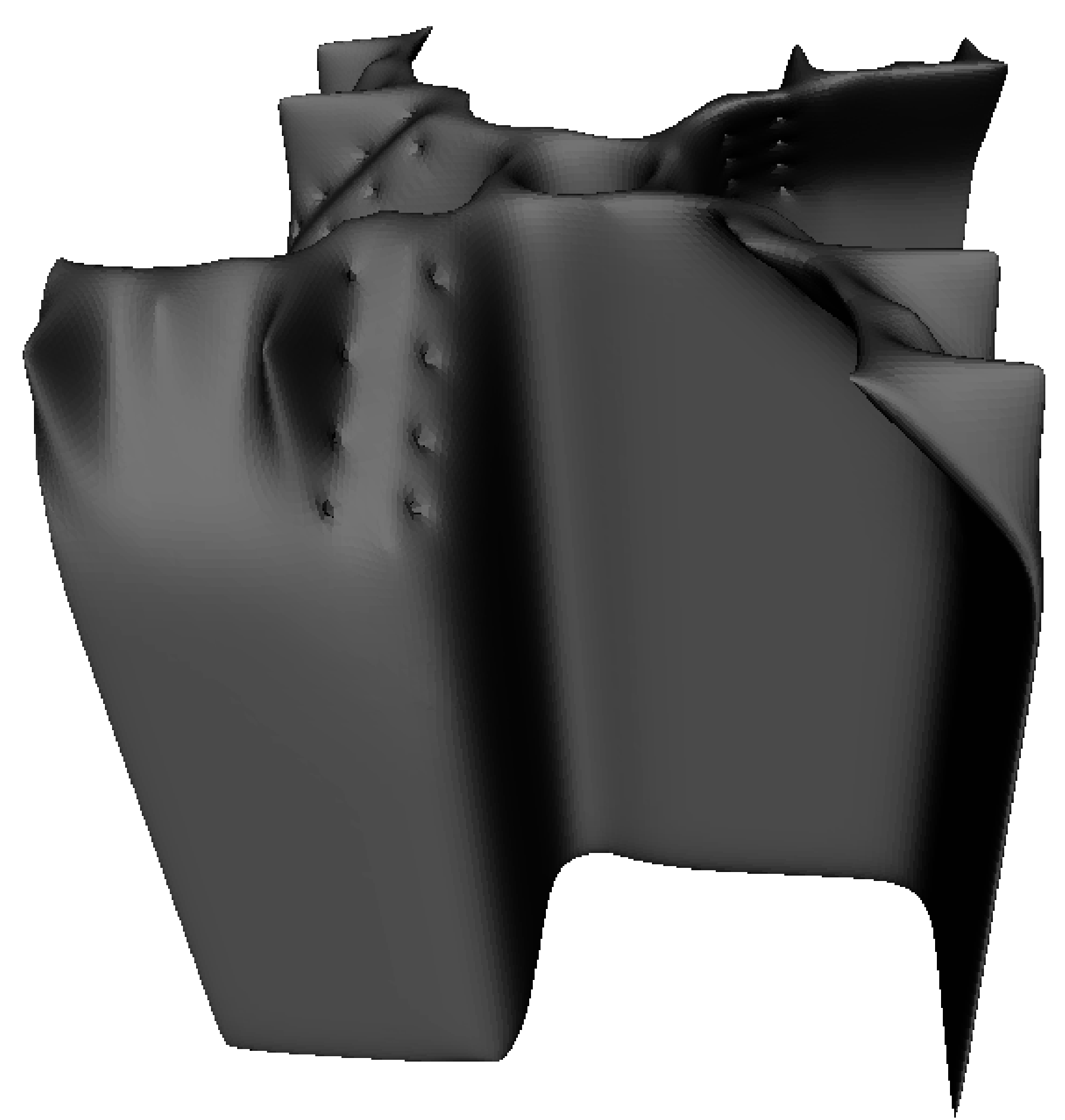}}
	\subfigure[5th Gen]{\label{fig:zn-g5-top}\includegraphics[width=\zsmallfigwidth]{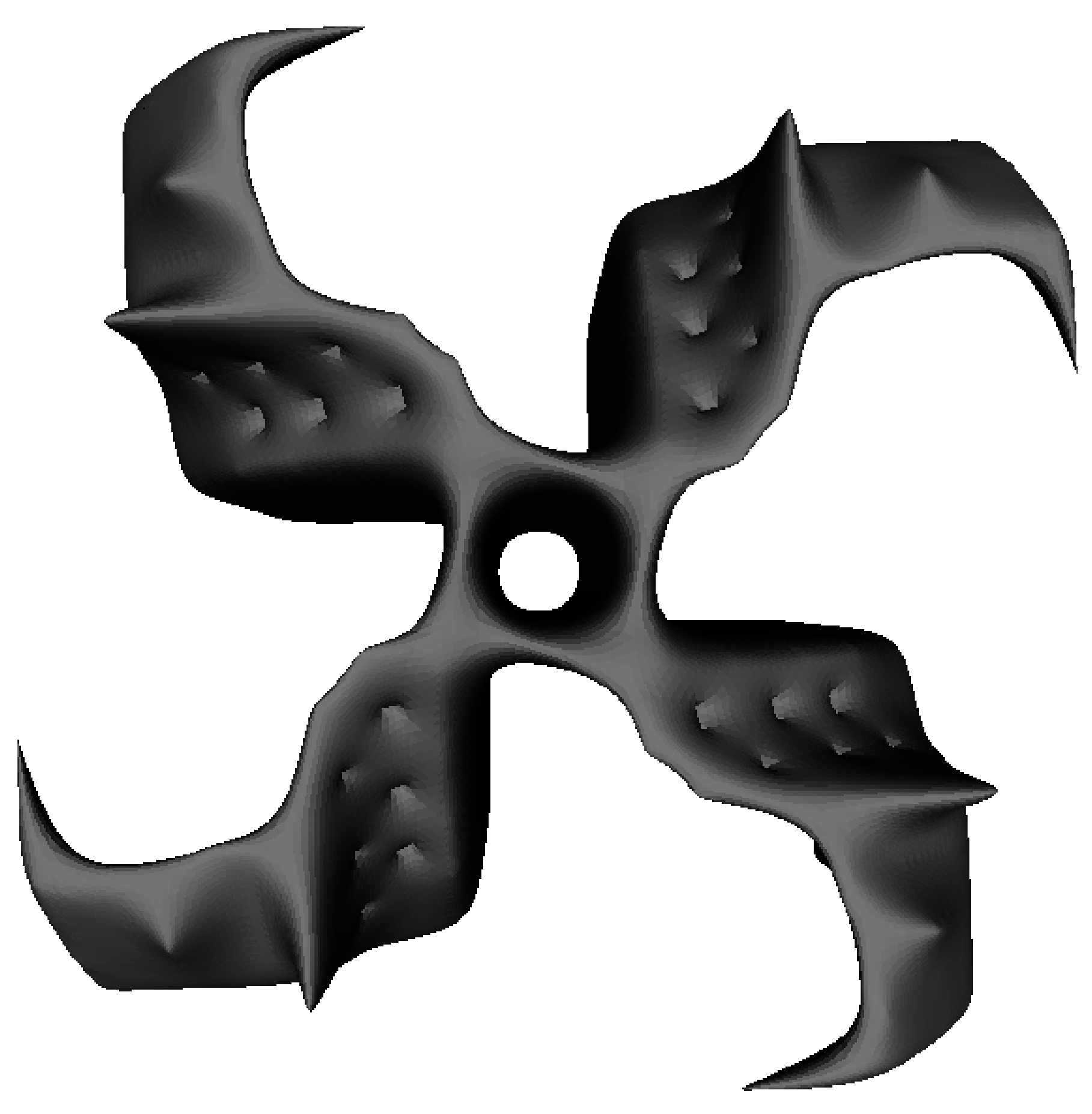} \includegraphics[width=\zsmallfigwidth]{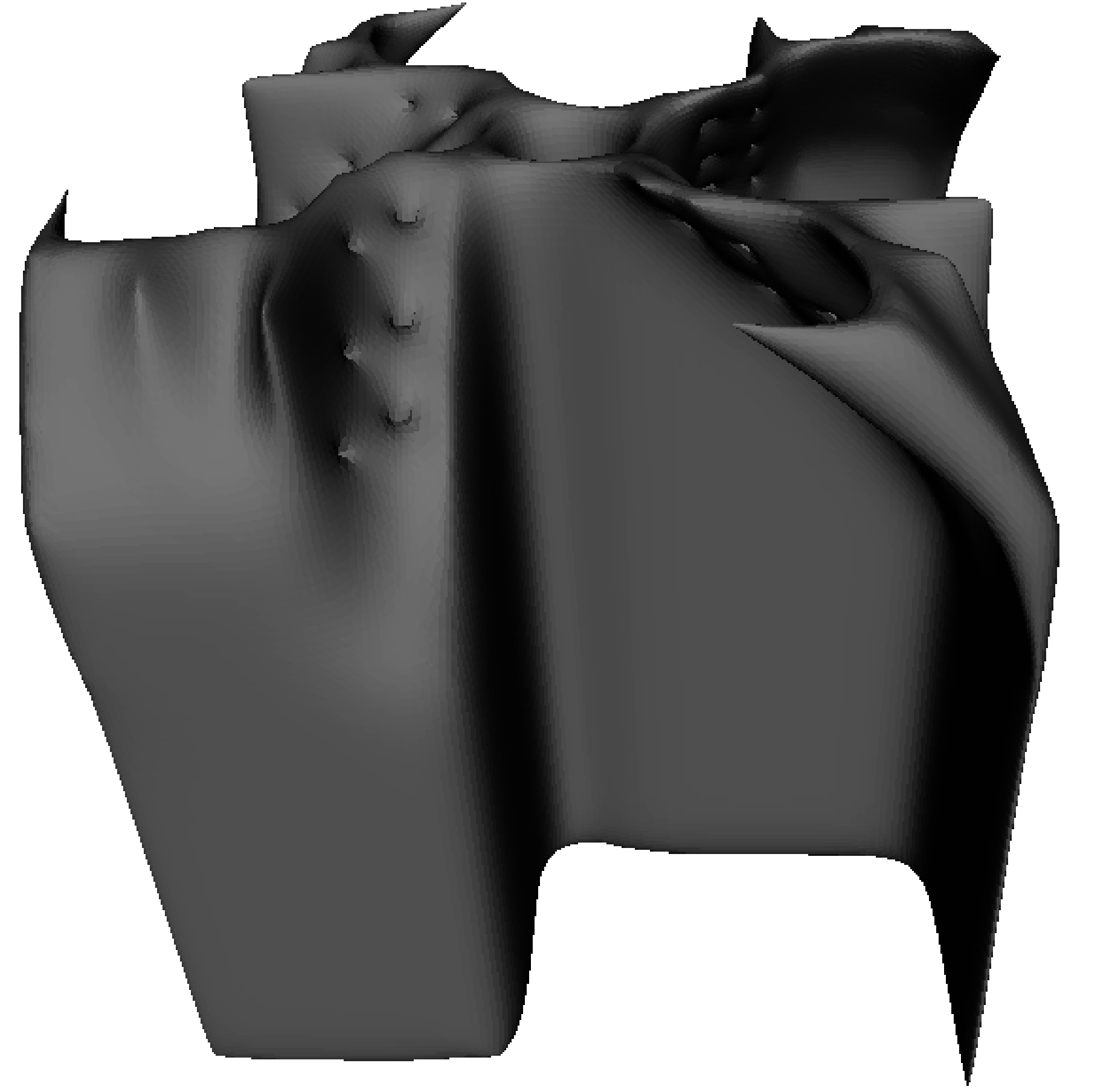}}
	\caption{The fittest evolved individuals with $z$-variability produced by the SGA each generation.}
	\label{fig:zmodel-individuals}
\end{figure}

Fig.~\ref{fig:zreal-res} shows the maximum rotation speed achieved each generation by the fittest individuals with $z$-variability. As before, 20 random designs are generated, fabricated, and evaluated to form the initial population and the GA is run for 2 further generations before the SGA is used. The fittest individuals produced by the GA and SGA each generation are shown in Figs.~\ref{fig:zga-individuals} and~\ref{fig:zmodel-individuals}, respectively. Again, both the GA-only and model-assisted approach design increasingly efficient prototypes. Comparing the final 10 individuals from each experiment, the average rotation speed of the SGA ($M=1217$, $SD=78$, $N=10$) is significantly greater than the GA-only approach ($M=1110$, $SD=41$, $N=10$) using a two-sample {\it t\/}-test assuming unequal variances, $t(14)=2.14$, $p\le.0018$. Furthermore, the fittest individual designed by the SGA ($1308rpm$) was greater than the GA-only approach ($1245rpm$) after 100 fabrications. The addition of the extra degree of freedom on the $z$-axis resulted in improved performance for both GA and model-assisted approaches ({\it cf.\/} Fig.~\ref{fig:real-res}).


\begin{figure}[t]
	\centering 
	\subfigure{\includegraphics[width=\zsmallfigwidth]{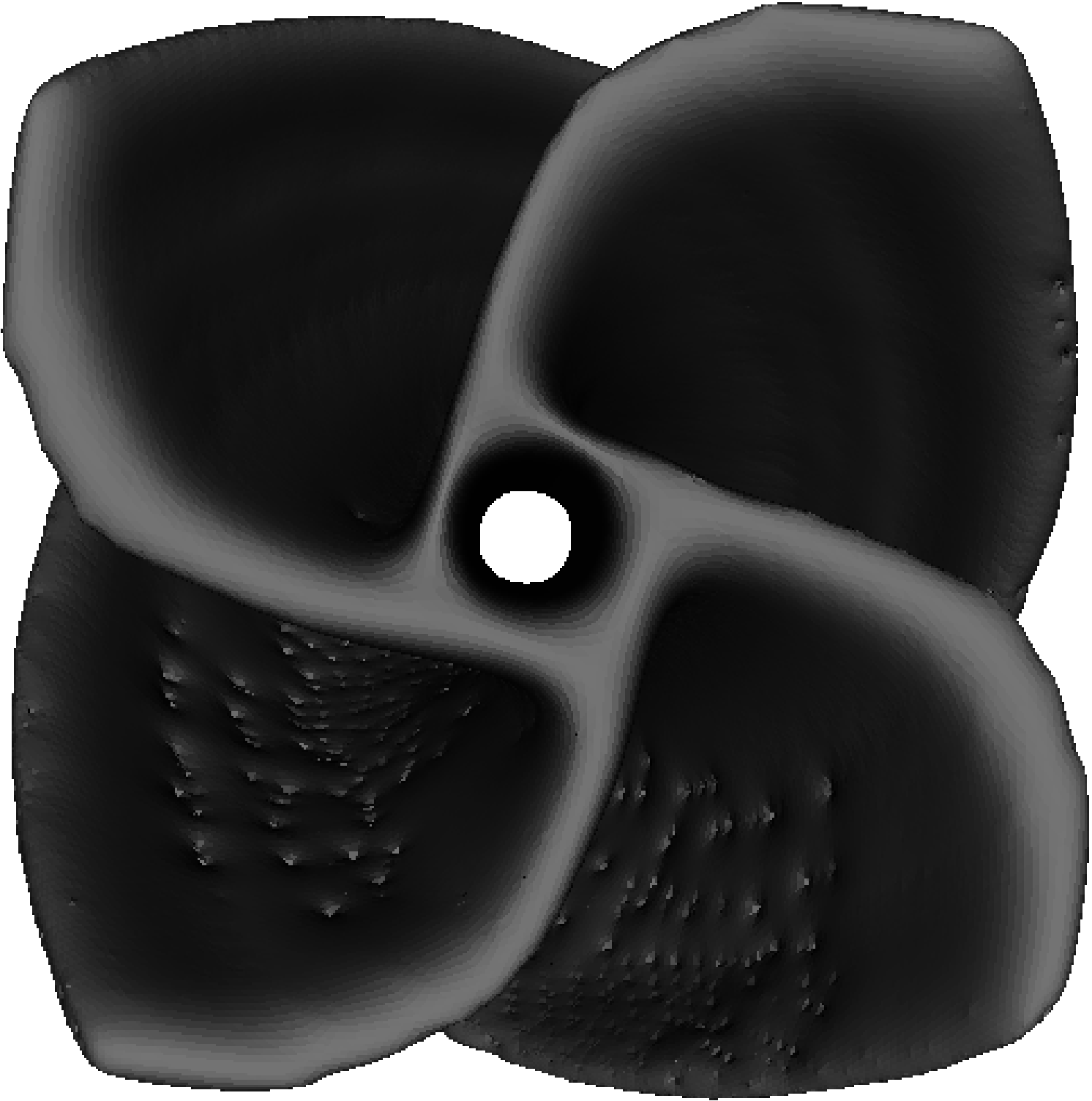}} \hspace{-0.1in}
	\subfigure{\includegraphics[width=\zsmallfigwidth]{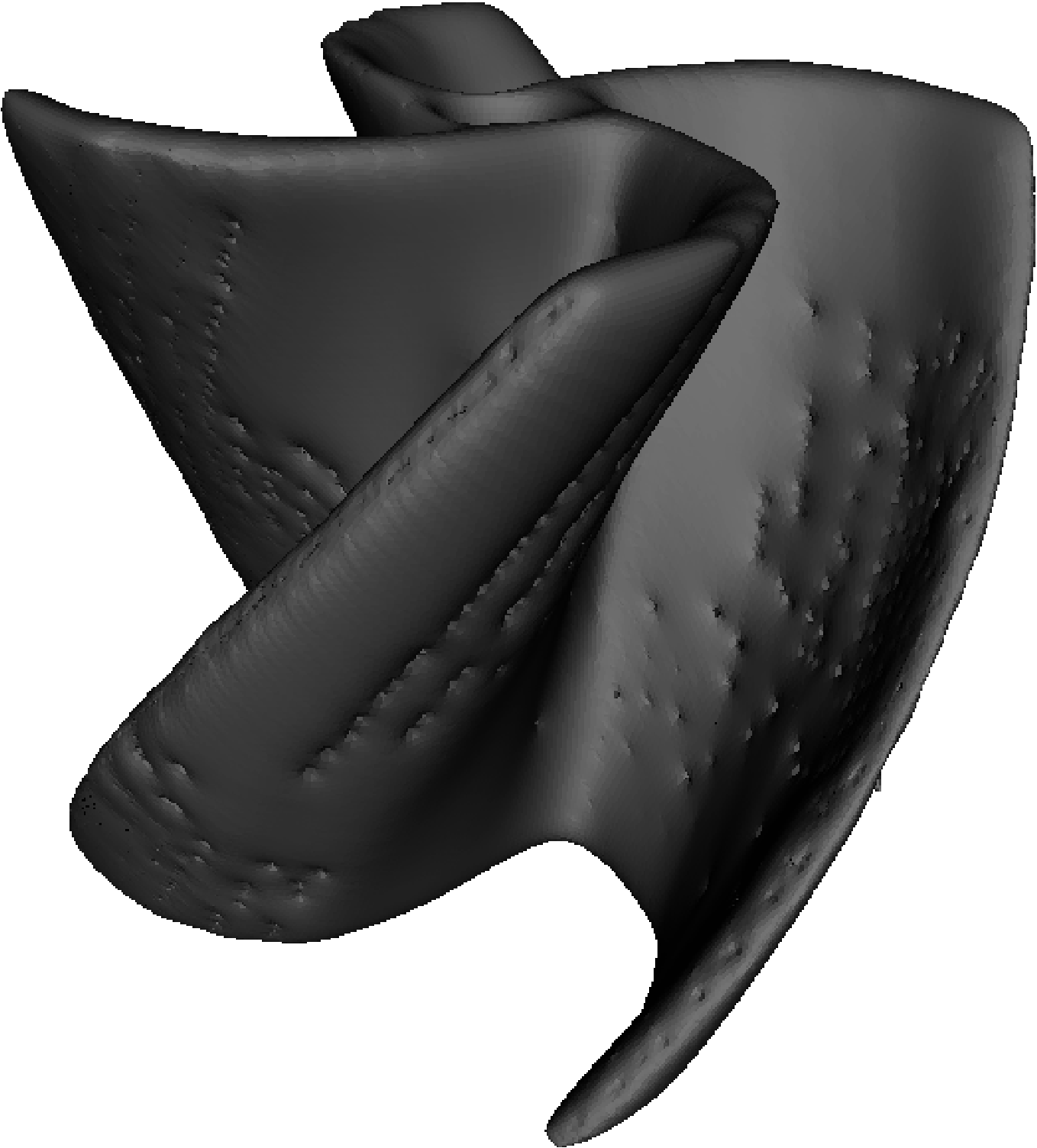}}
	\caption{Example from Fig~\ref{fig:target-smoothed} $z$-rotated (as in U.S. patent 7,371,135)}\label{fig:patent}
\end{figure}

\begin{figure}[t]
	\centering 
	\subfigure{\includegraphics[width=\zsmallfigwidth]{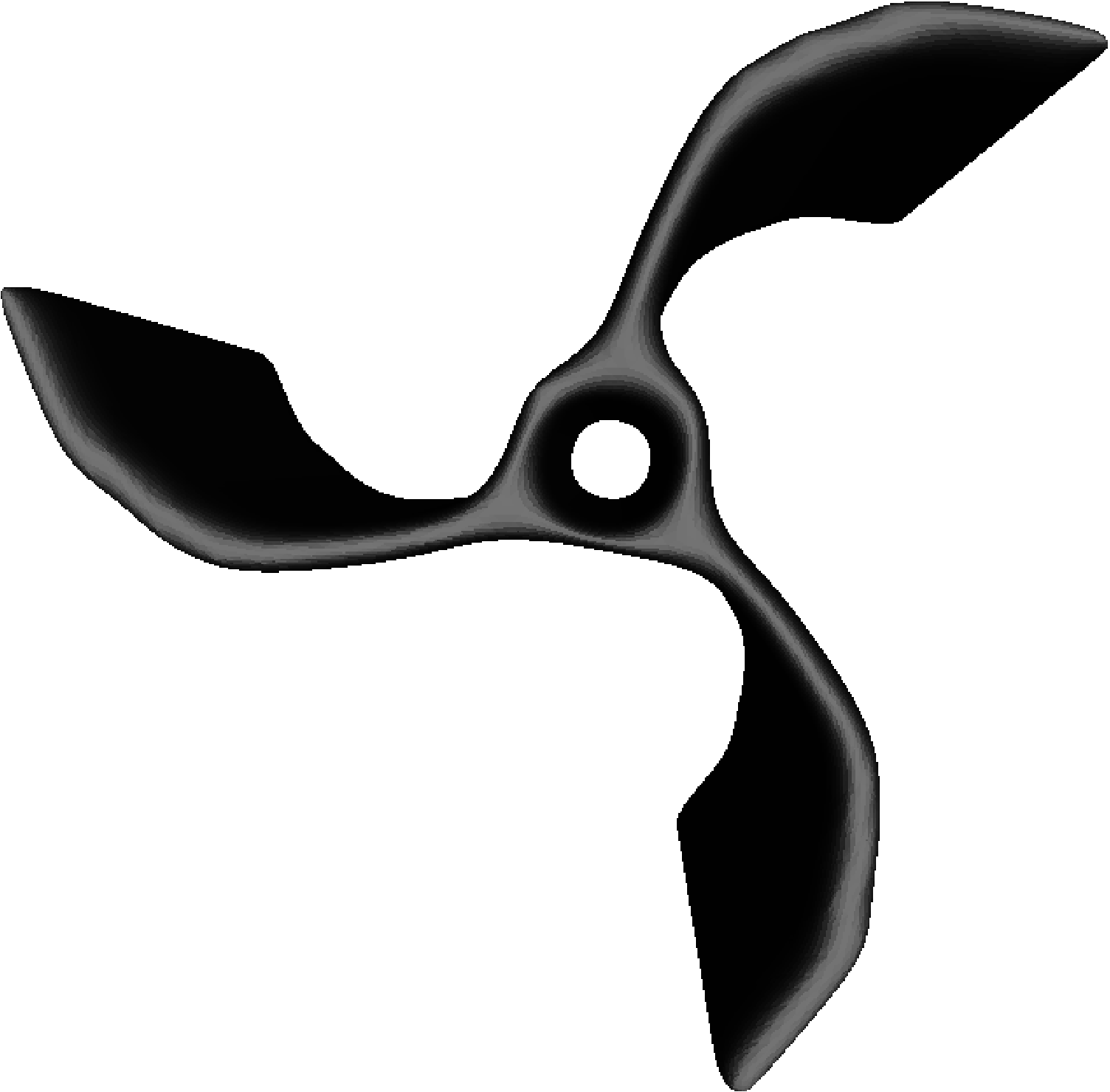}} \hspace{-0.1in}
	\subfigure{\includegraphics[width=\zsmallfigwidth]{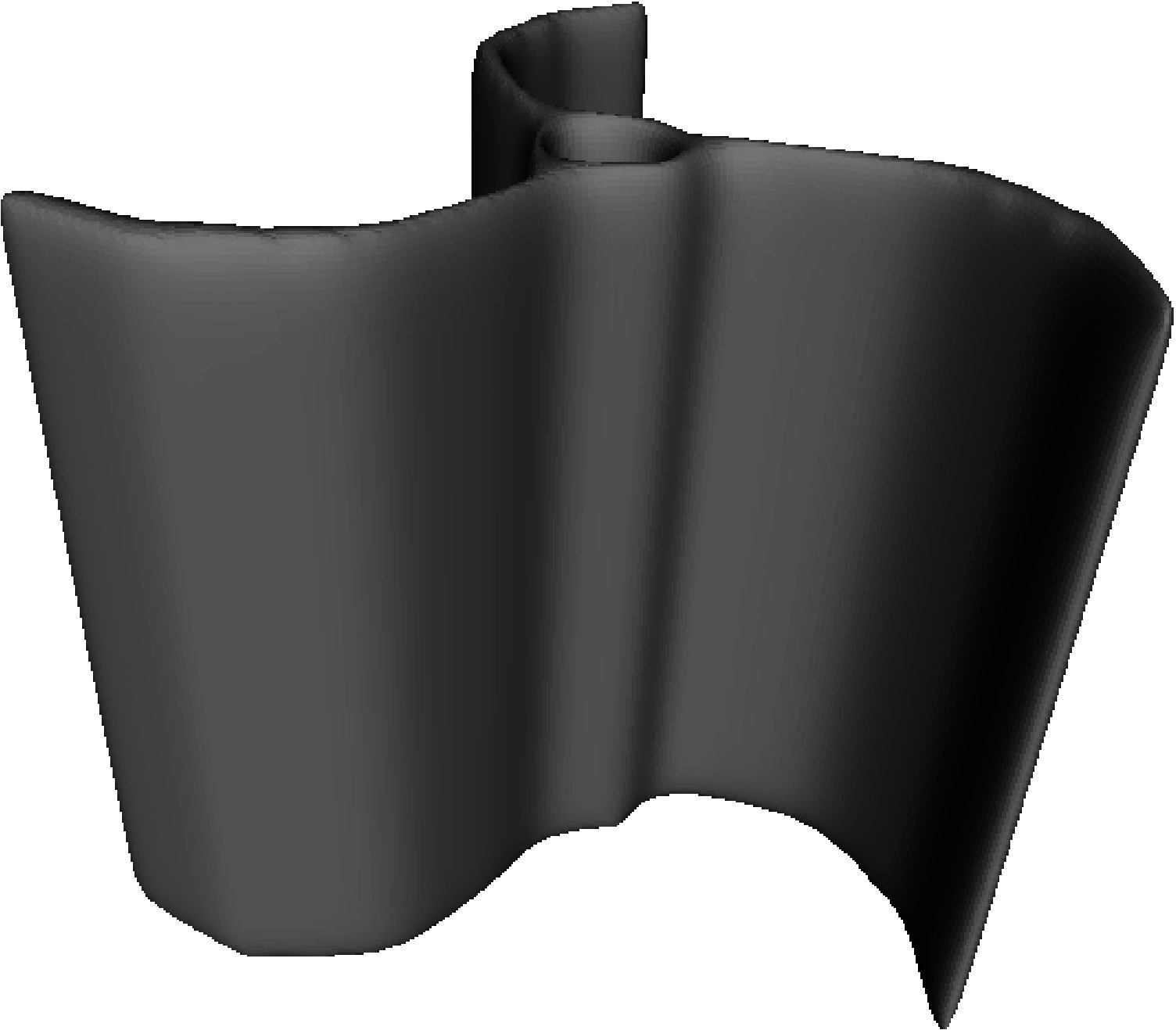}}
	\caption{Classic 3 blade Savonius}\label{fig:3blade}
\end{figure}

\section{VAWT Array Rotation-Based Evolution}
\label{sec:array}
     
Additional challenges are encountered when extracting large amounts of wind energy since multiple turbines must be arranged into a wind farm. As the turbines extract the energy from the wind, the energy content decreases and the amount of turbulence increases downstream from each. See~\cite{Hasager:2013} for photographs and explanation of the well-known wake effect at the Horns Rev offshore wind farm in the North Sea. Due to this, HAWTs must be spaced 3--5 turbine diameters apart in the cross-wind direction and 6--10 diameters apart in the downwind direction in order to maintain 90\% of the performance of isolated HAWTs~\cite{Dabiri:2011}. The study of these wake effects is therefore a very complex and important area of research~\cite{Barthelmie:2006}, as is turbine placement~\cite{Mosetti:1994}. However, the spacing constraints of HAWT often do not apply for VAWT, and VAWT performance can even be increased by the exploitation of inter-turbine flow effects~\cite{Charwat:1978}. Indeed, it has recently been shown~\cite{Dabiri:2011} that power densities an order of magnitude greater can be potentially achieved by arranging VAWTs in layouts utilising counter-rotation that enable them to extract energy from adjacent wakes and from above the wind farm.

The use of approximations in a coevolutionary context has previously been shown capable of solving computationally expensive optimisation problems with varying degrees of epistasis more efficiently than conventional CGAs through the use of radial basis functions~\cite{Ong:2002} and memetic algorithms~\cite{Goh:2011}.
 
\begin{figure}[t]
	\centering 
	\subfigure{\includegraphics[width=3.0in]{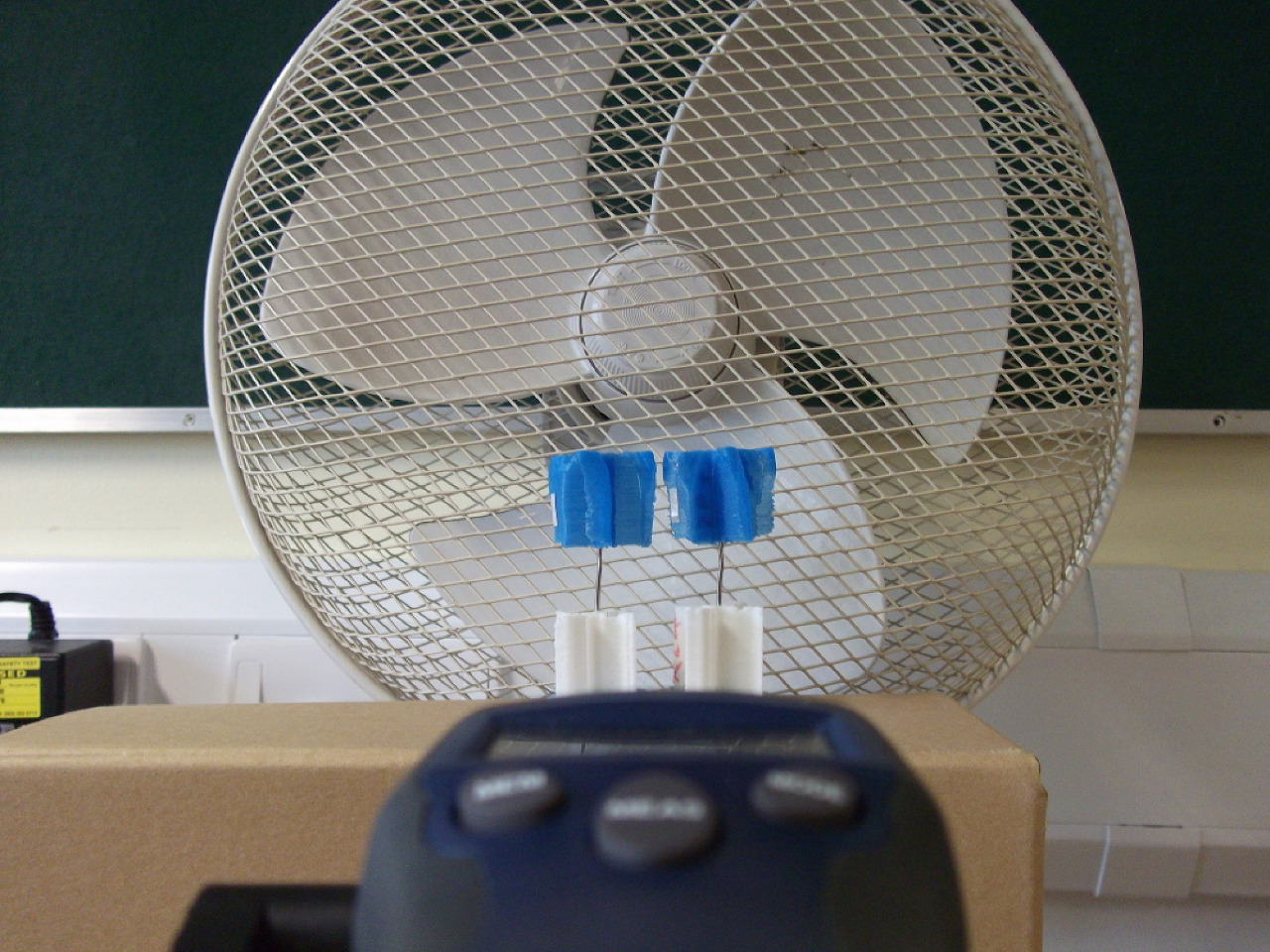} }
	\caption{VAWT array experimental setup.}
	\label{fig:setup-double}
\end{figure}

Here, we investigate a surrogate-assisted cooperative coevolutionary approach (SCGA) to design wind farms, utilising the aggregated rotation speed of the array as fitness. Each VAWT is treated separately by evolution and approximation techniques, i.e.,\ heterogeneous designs could emerge. In addition, a Boolean gene is added to designate the rotation of a turbine in order to allow counter-rotating arrays to potentially emerge. Two turbines are positioned $33mm$ adjacently and $30mm$ from the propeller fan. That is, there is a $3mm$ spacing between the blades at their closest point; see the experimental setup for two closely positioned VAWT in Fig~\ref{fig:setup-double}.

The individuals in each species population are initially evaluated in collaboration with a single randomly selected individual from the other species population. Thereafter, the GA proceeds as before, however alternating between species after each offspring is formed and evaluated with the elite member from the other species; see algorithm outline in Algorithm~\ref{alg:cga} and \cite{Bull:1997b} for discussions on collaboration strategies. In the SCGA, the models use identical parameters to the single VAWT experiments, however 16 input neurons are now required. In addition, the model weights must be reinitialised each time before training due to the temporal nature of pairing with the elite member, and the GA runs for one generation (using the model approximated fitnesses where real fitness is unknown) before the individual with the highest approximated fitness and a randomly selected unevaluated individual are evaluated with the elite member from the other species; see outline in Algorithm~\ref{alg:scga}.

\begin{algorithm}[t]
	\SetAlgoLined%
	Generate and fabricate individuals for all species\;
	\For{each species population}{
		Select random representative from each species\;
		\For{each individual in population}{
			Evaluate\;
		}
	}
	\While{fabrication budget not exhausted}{
		\For{each species population}{
			Create an offspring using evolutionary operators\;
			Select representatives for each species\;
			Fabricate and evaluate the offspring\;
			Add offspring to species population\;
		}
	}
	\caption{Coevolutionary genetic algorithm}
	\label{alg:cga}
\end{algorithm}

\begin{algorithm}[t]
	\SetAlgoLined%
	Generate and fabricate individuals for all species\;
	\For{each species population}{
		Select random representative from each species\;
		\For{each individual in population}{
			Evaluate\;
			Add individual to species evaluated list\;
		}
	}
	\While{fabrication budget not exhausted}{
		\For{each species population}{
			Initialise model weights\;
			Train model on species evaluated list\;
			\For{each individual in population}{
				\If{individual unevaluated}{
					Set approximated fitness\;
				}
			}
			\For{population size number of times}{
				Create offspring using evolutionary operators\;
				Set offspring approximated fitness\;
				Add offspring to species population\;
			}
			Select representatives for each species\;
			Fabricate, evaluate, and add to species evaluated list, the individual with the highest approximated fitness in species\;
			Fabricate, evaluate, and add to species evaluated list, a random unevaluated individual in species\;
		}
	}
	\caption{Surrogate-assisted coevolutionary algorithm}
	\label{alg:scga}
\end{algorithm}
 
To help increase performance and save fabrication time, each species population in the VAWT array experiment initially consists of the first 10 ($>0rpm$) individuals from the previous single VAWT ($z$-varying) experiment, both normally rotated and counter-rotated. These individuals still retain a good degree of randomness while possessing some useful aerodynamic properties. Each species thus maintains a population of 20 individuals.
                                                                                                                                                                                                            
\begin{figure}[t]
	\centering 
	\includegraphics[width=\graphwidth]{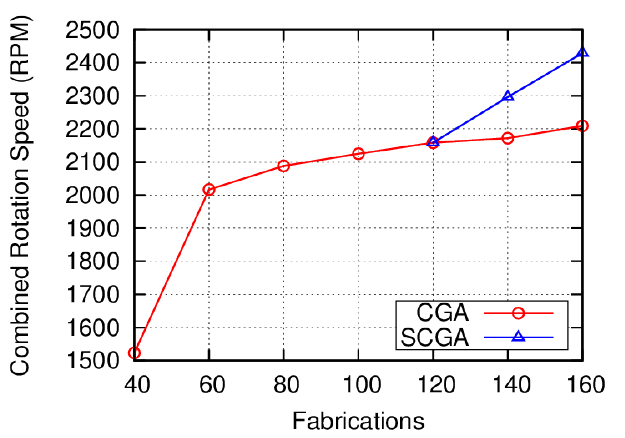}
\caption{Array rotation-based evolution. Fittest CGA (circle) and SCGA (triangle) arrays. The SCGA is used for comparison only after 120 evaluations (i.e., 3 generations) of the CGA since sufficient training data is required for the surrogate models.}
	\label{fig:array-best-res}
\end{figure}

\begin{figure}[t]
	\centering 
	\subfigure[1st Gen]{ \includegraphics[width=\zsmallfigwidth]{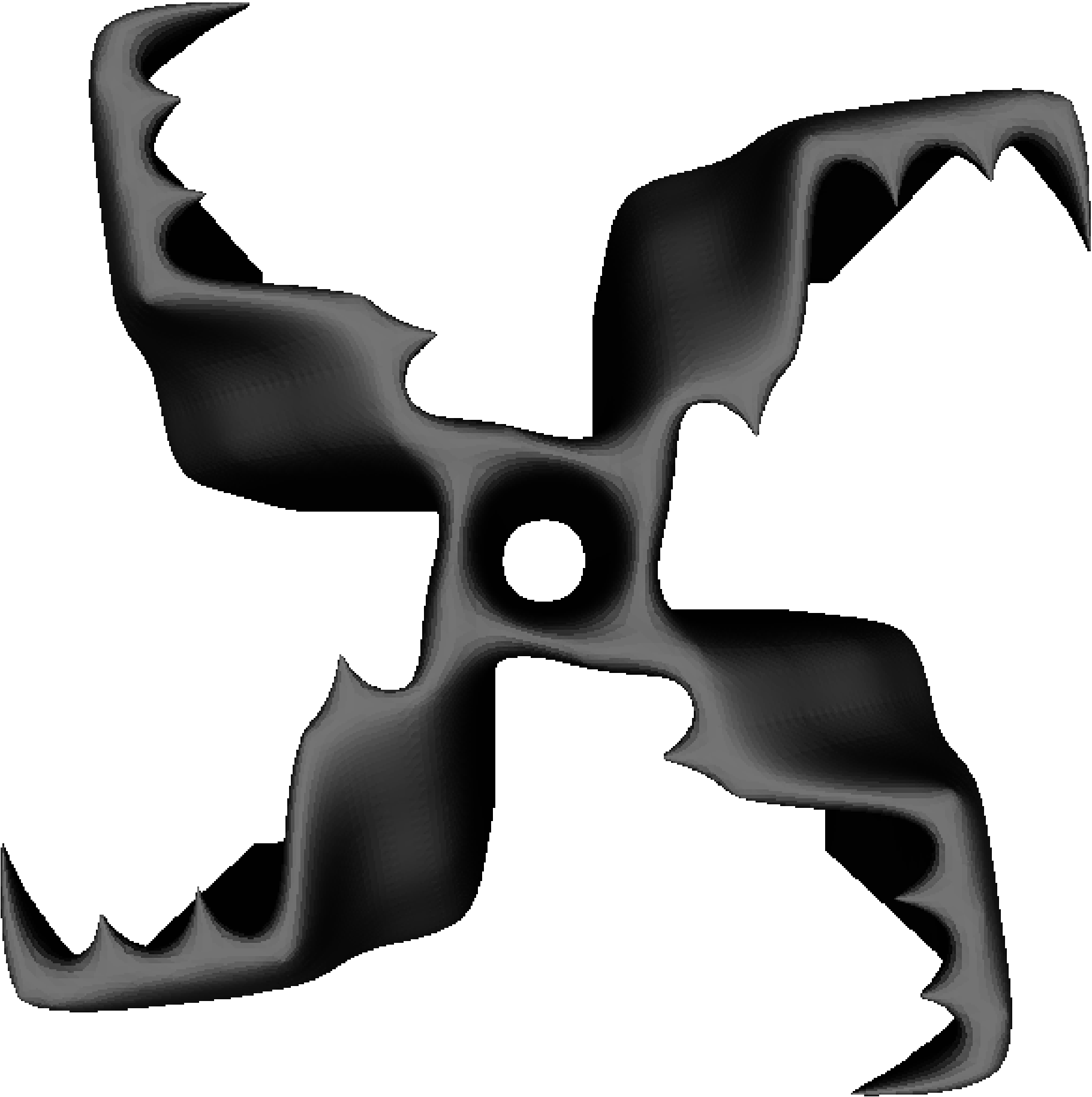}   \includegraphics[width=\zsmallfigwidth]{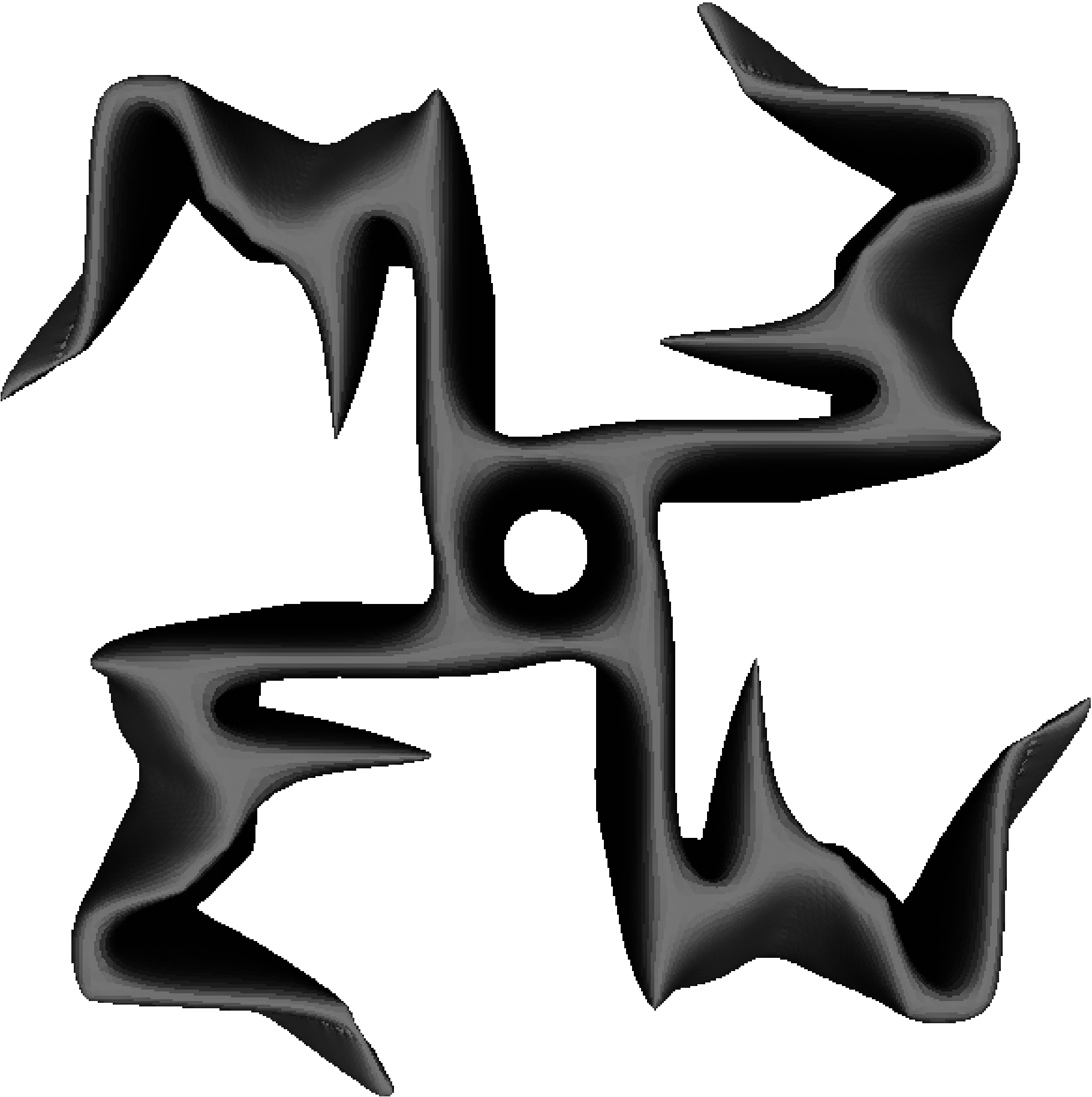}}
	\subfigure[2nd Gen]{ \includegraphics[width=\zsmallfigwidth]{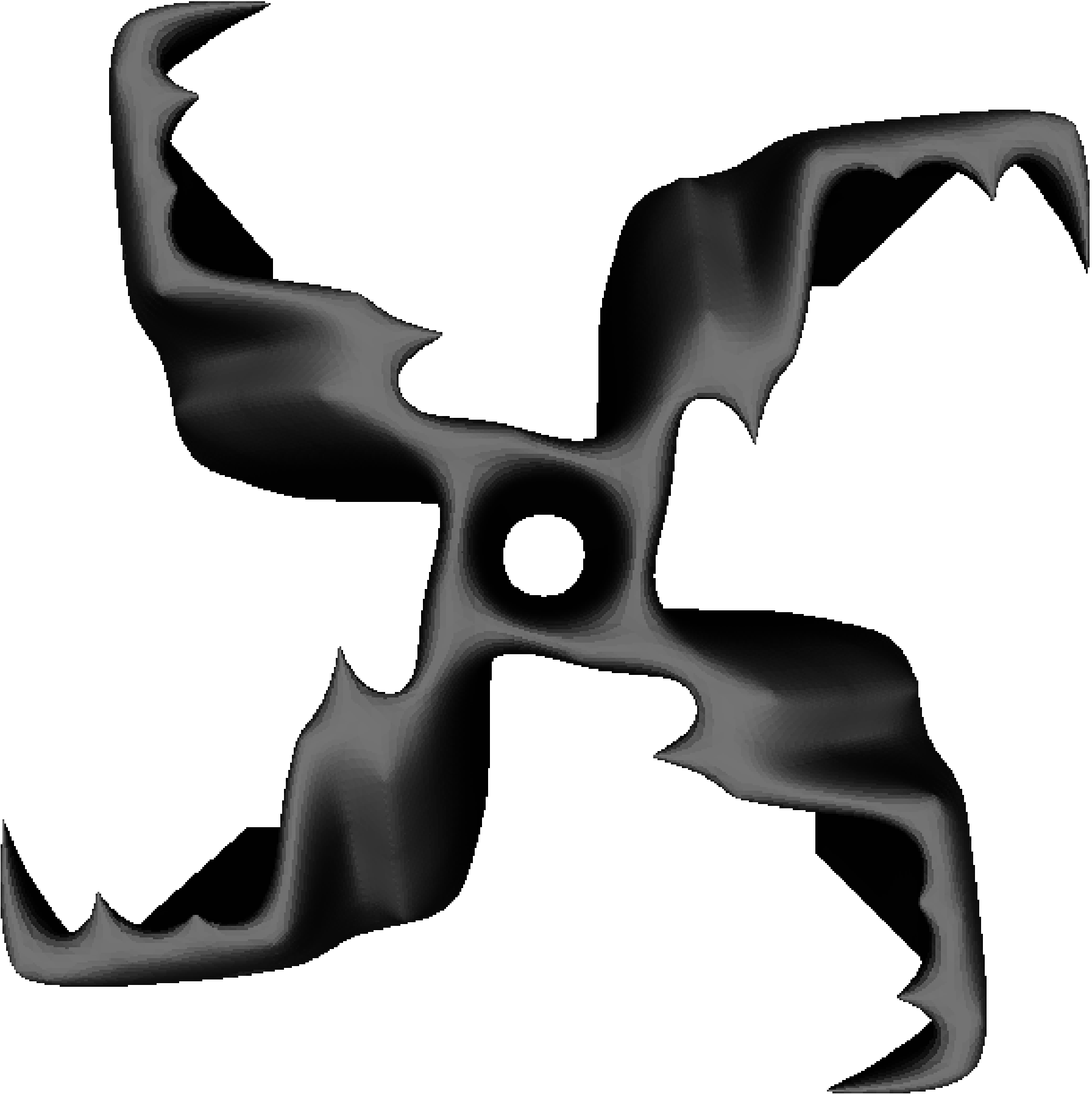} \includegraphics[width=\zsmallfigwidth]{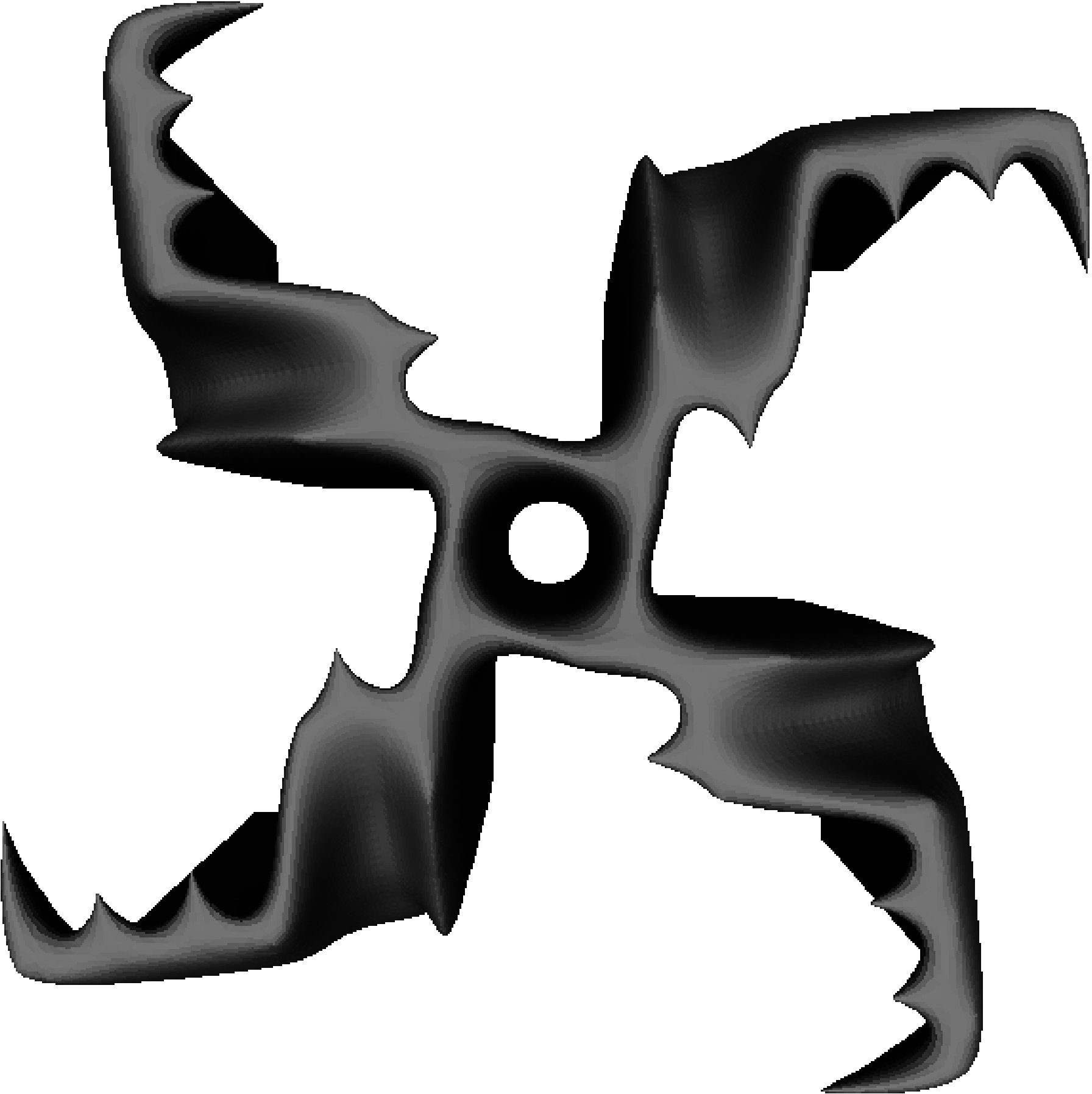}}
	\subfigure[3rd Gen]{ \includegraphics[width=\zsmallfigwidth]{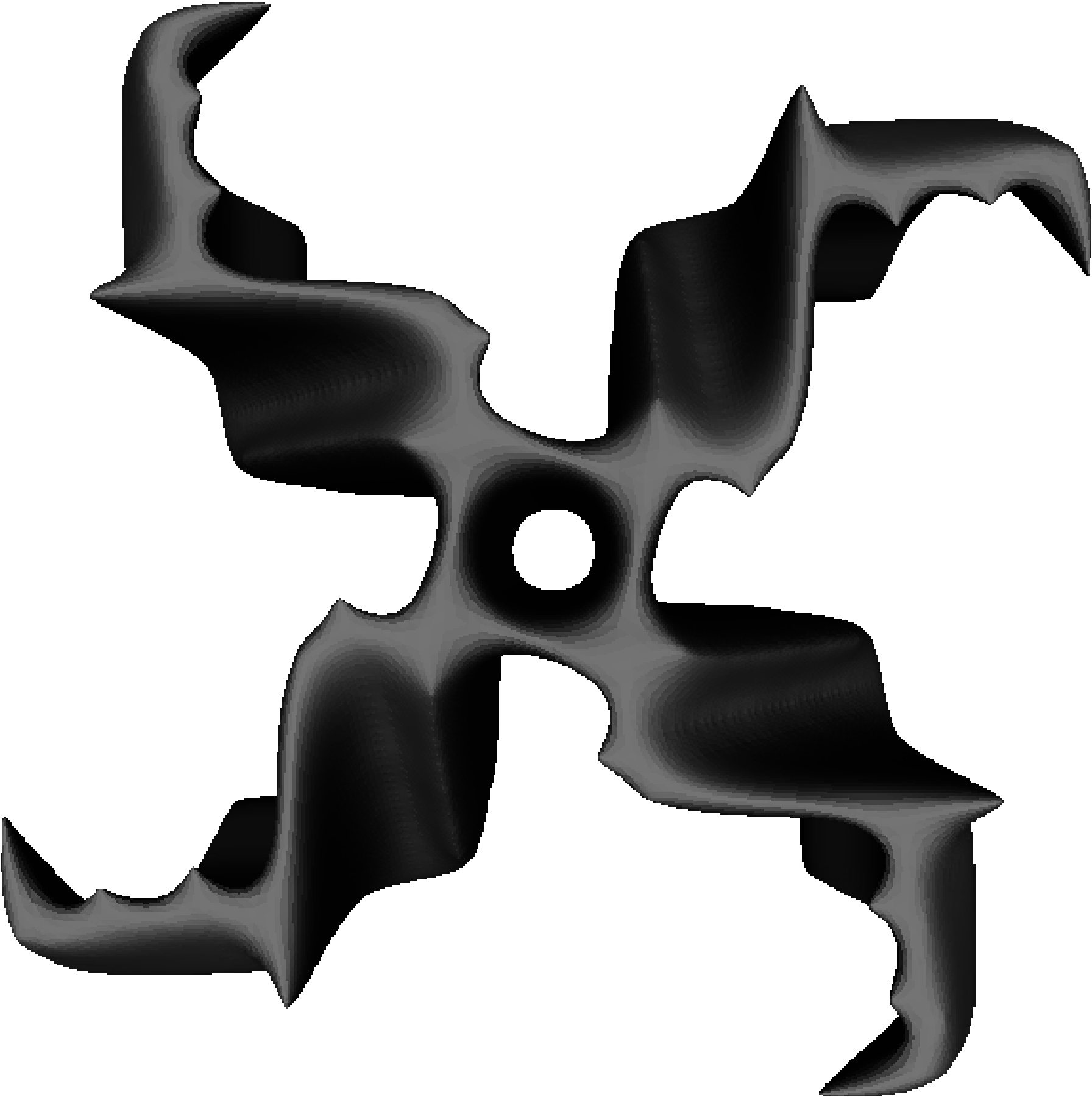} \includegraphics[width=\zsmallfigwidth]{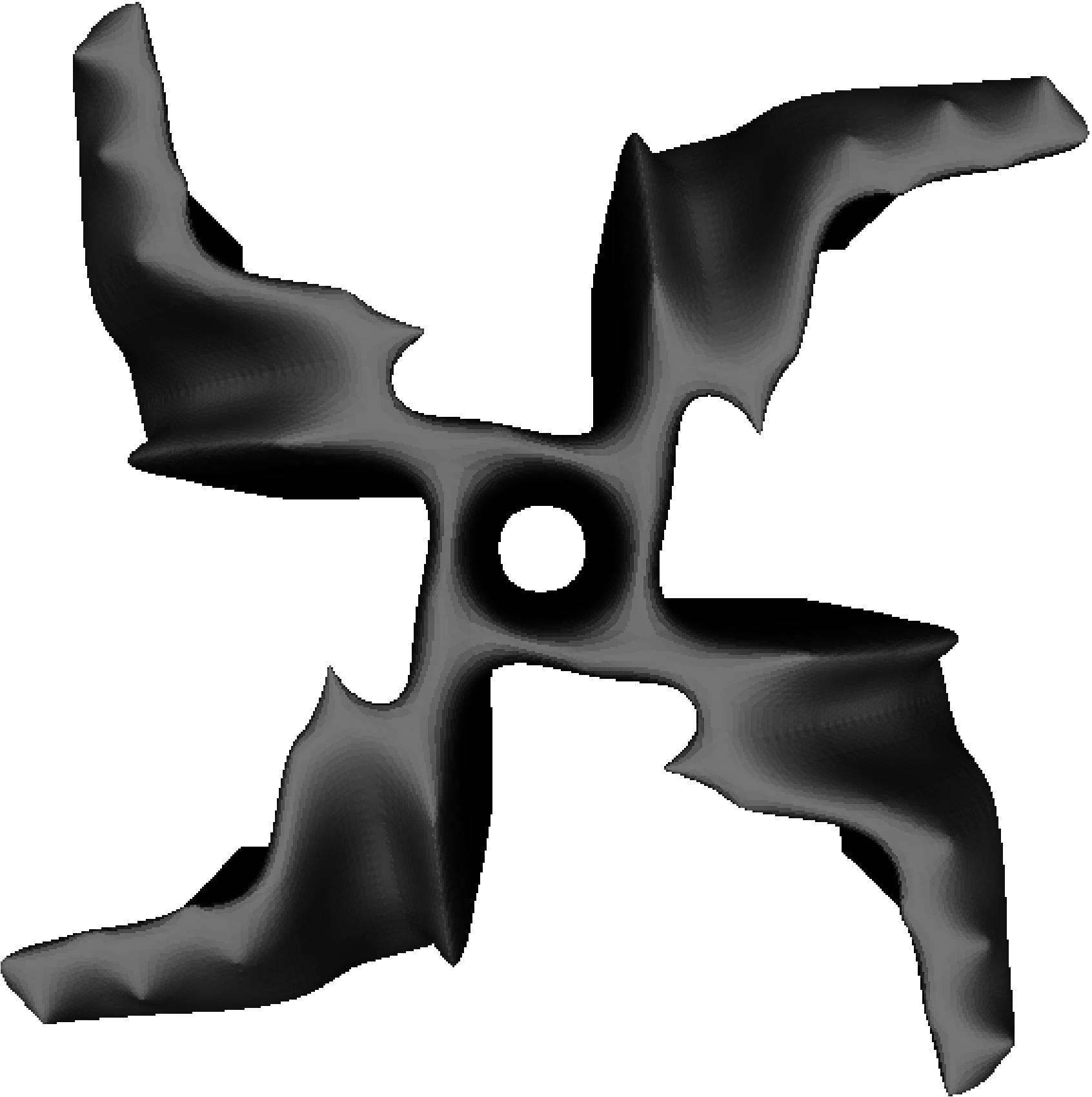}}
	\subfigure[4th Gen]{ \includegraphics[width=\zsmallfigwidth]{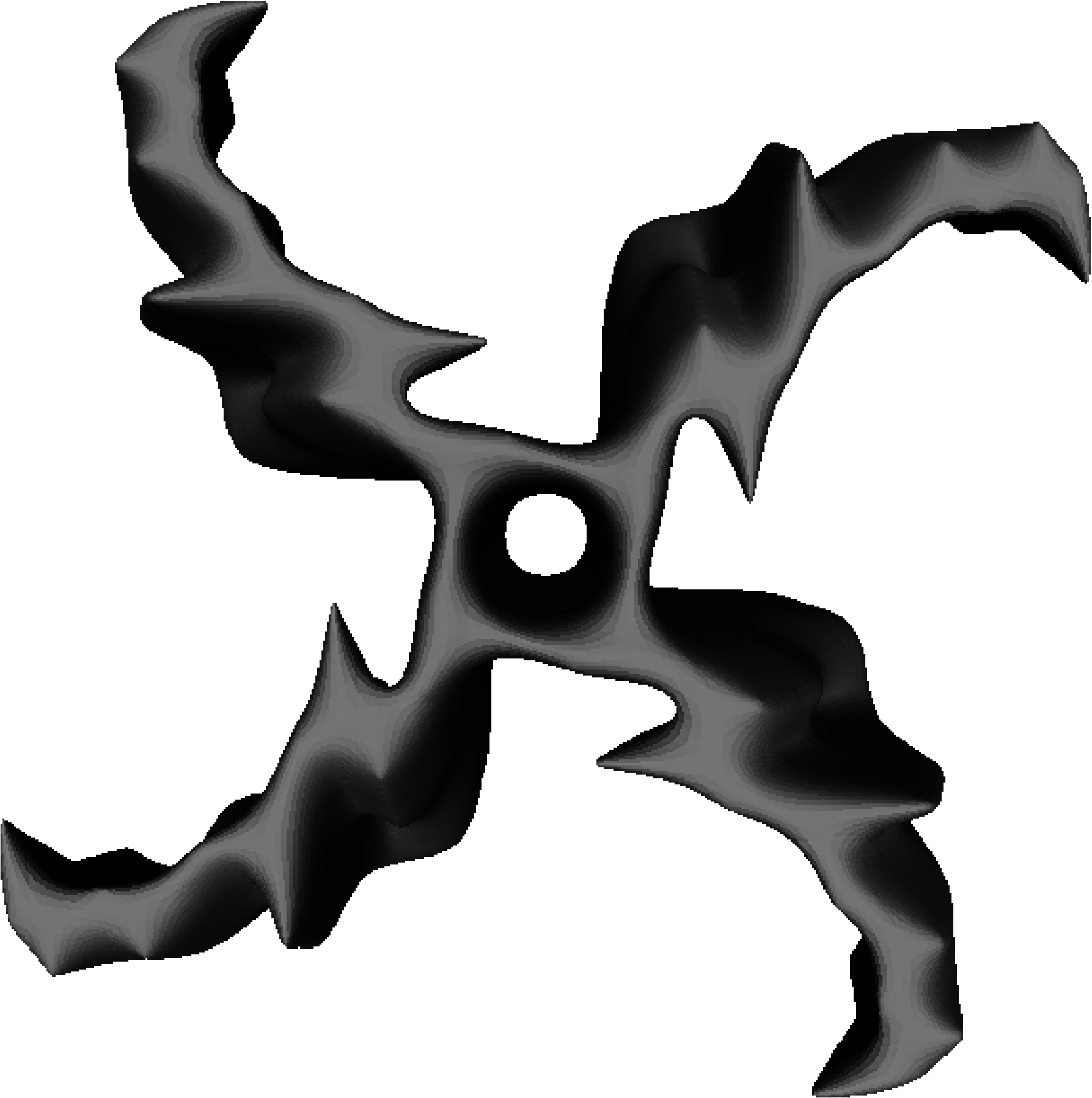} \includegraphics[width=\zsmallfigwidth]{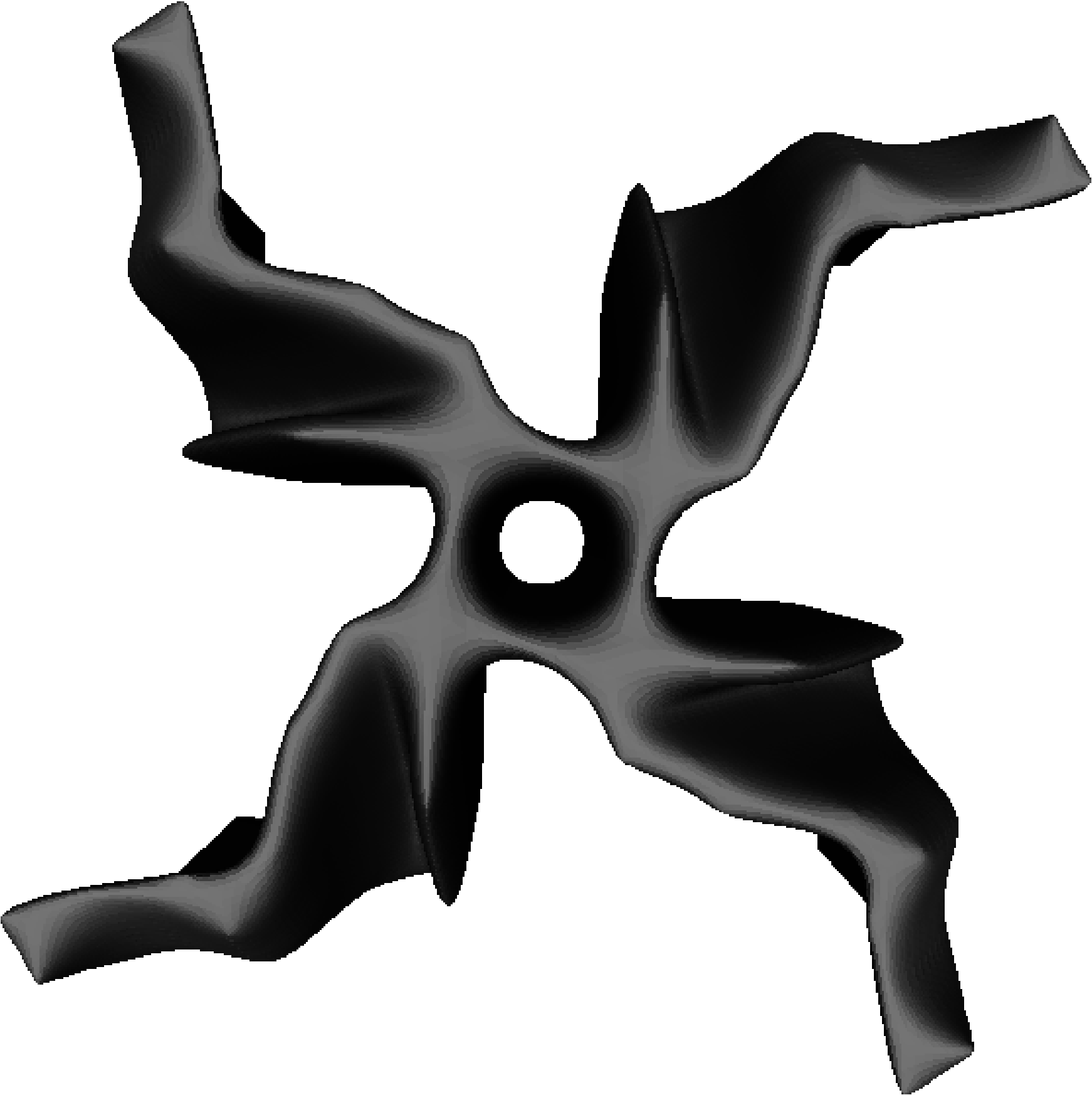}}
	\caption{Top view of the fittest CGA evolved heterogeneous array each generation. Wind direction from the south.}
	\label{fig:cga-individuals}
\end{figure}

\begin{figure}[t]
	\centering 
	\subfigure[4th Gen --- top view]{  \includegraphics[width=\zsmallfigwidth]{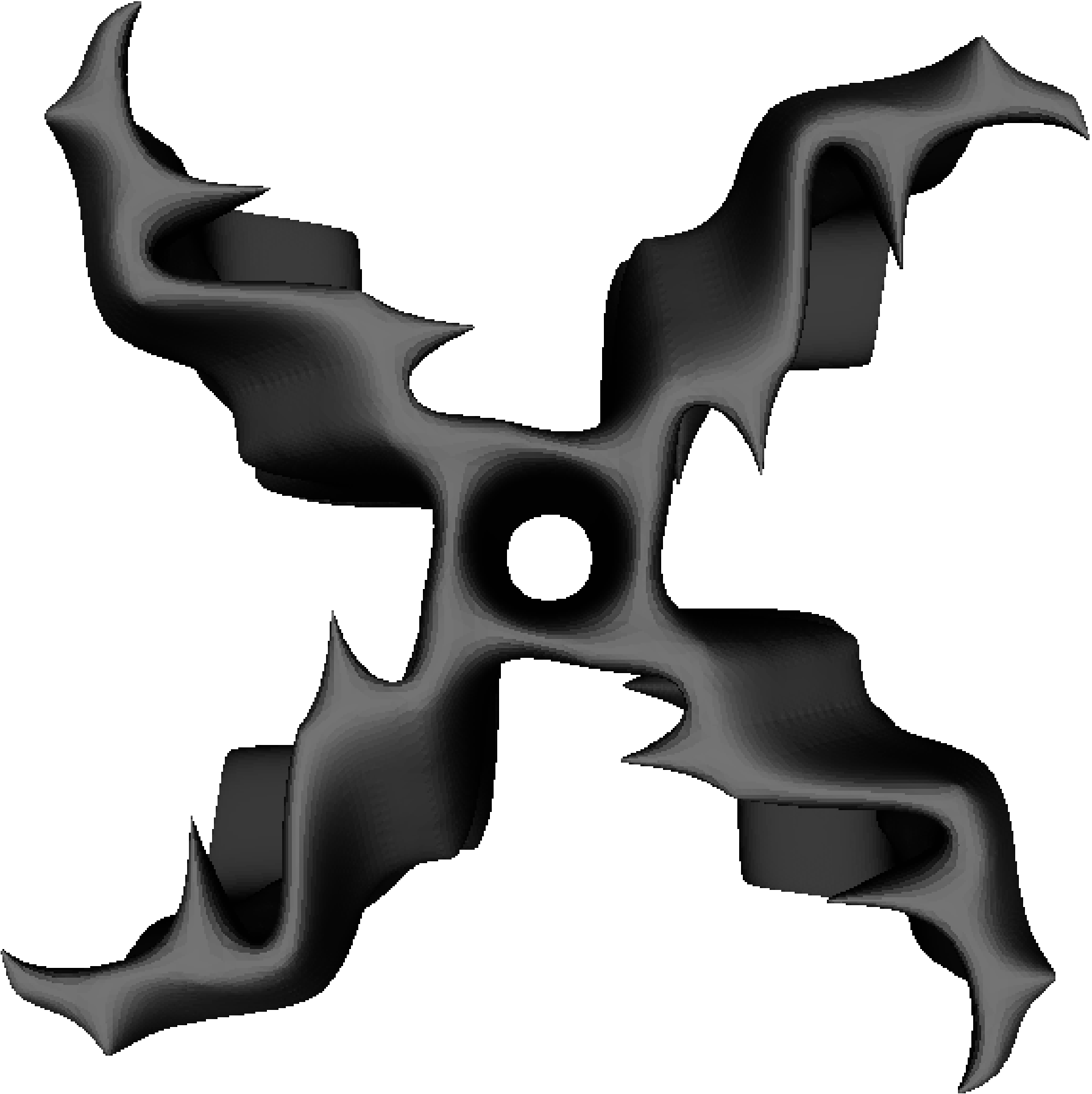} \includegraphics[width=\zsmallfigwidth]{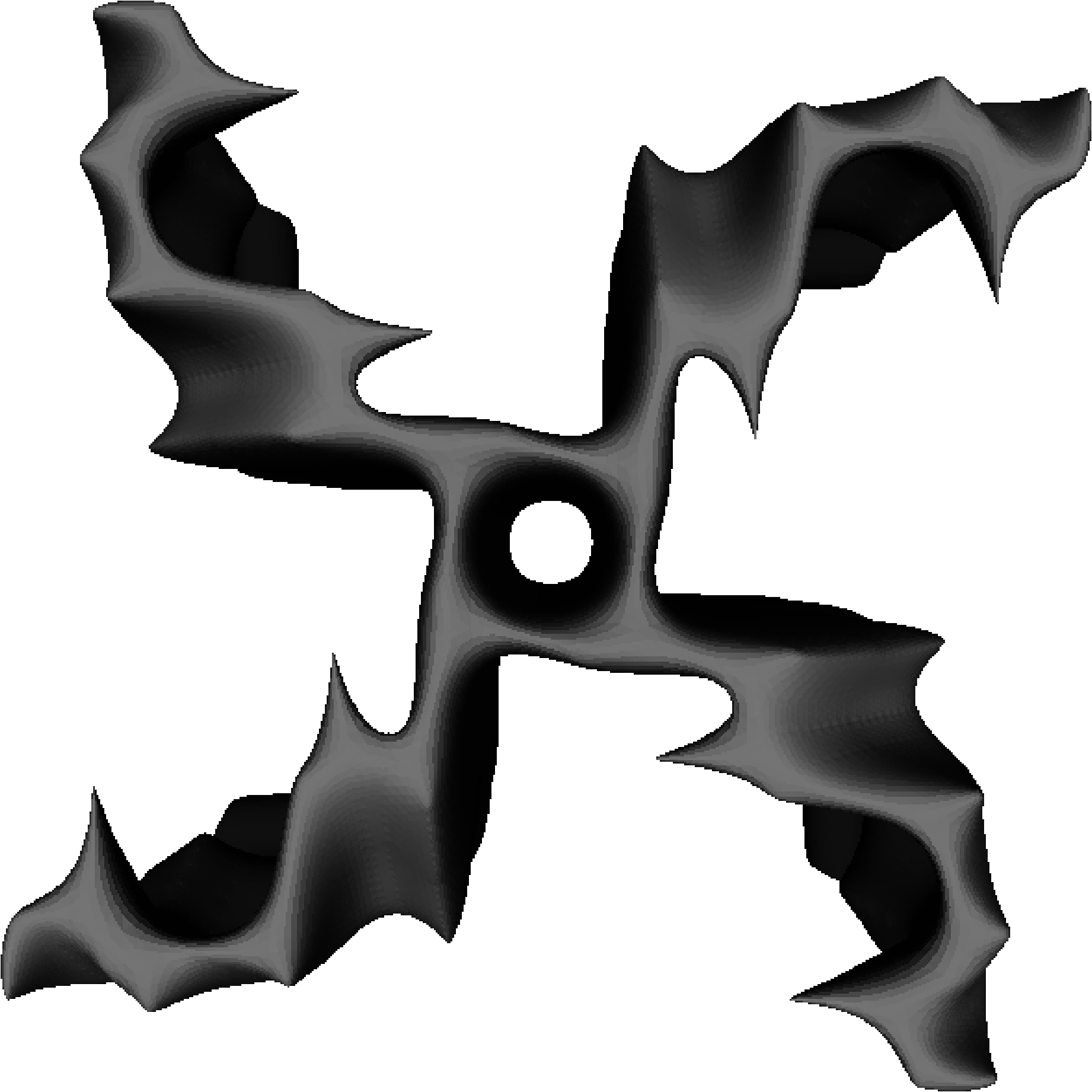}}
	\subfigure[4th Gen --- side view]{ \includegraphics[width=\zsmallfigwidth]{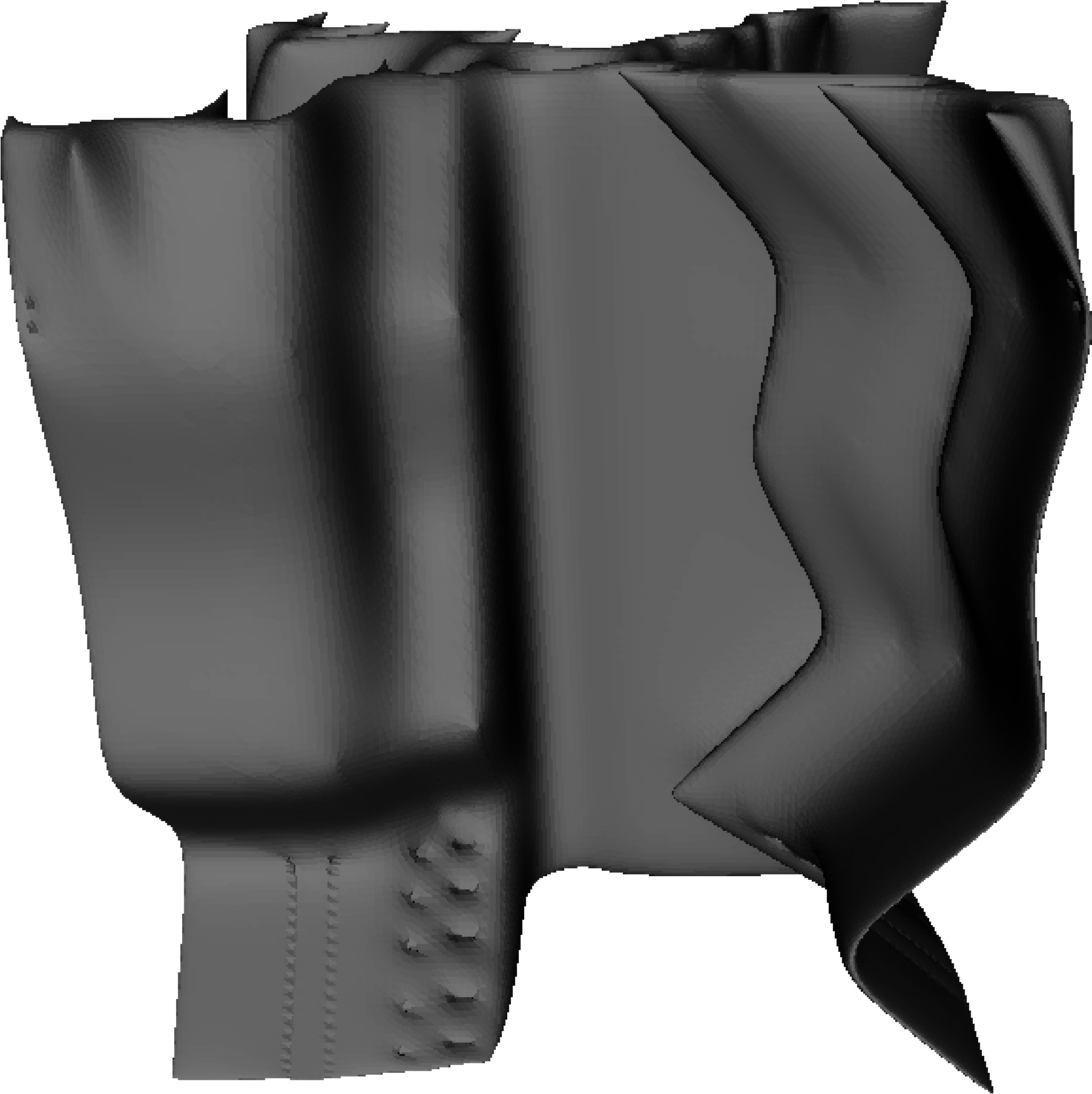} \includegraphics[width=\zsmallfigwidth]{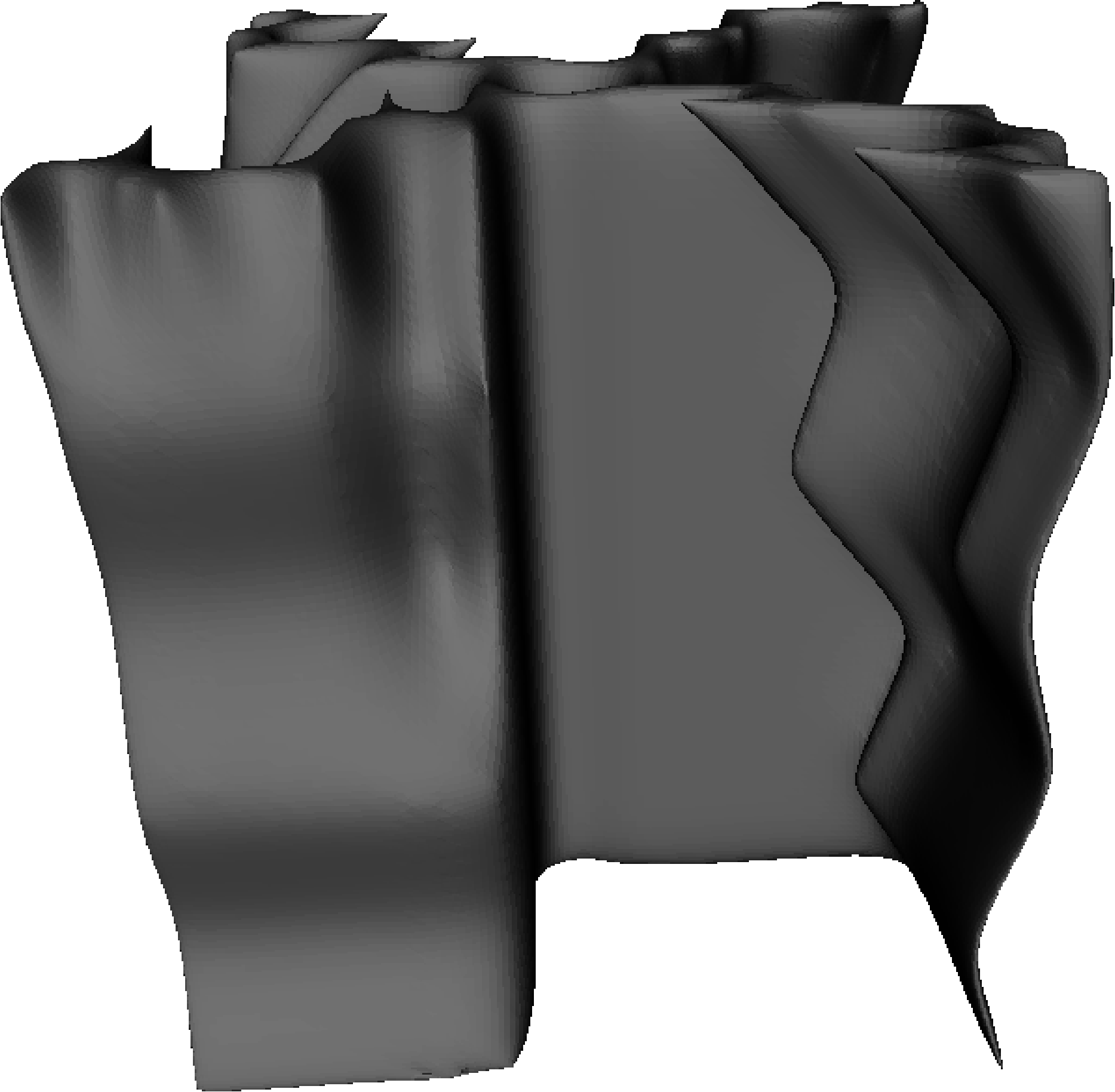}}
	\caption{The fittest SCGA evolved heterogeneous array. Wind direction from the south.}
	\label{fig:scga-individuals}
\end{figure}

The CGA and SCGA system performances are shown in Fig.~\ref{fig:array-best-res}, which illustrates the combined maximum rotation speed achieved by the fittest individuals each half-generation. As can be seen, the performance of the CGA initially increases rapidly from $1523rpm$ after 40 fabrications (that is, the initial random population) to $2088rpm$ after 80 fabrications (the second generation). This is largely due to the transition from initial random collaboration to pairing with the elite member in each species. Thereafter, the CGA continues to identify increasingly higher aggregated rotation speeds: $2158rpm$ after 120 fabrications, and $2209rpm$ after 160 fabrications (see array designs in Fig~\ref{fig:cga-individuals}). Similar to the previous experiments, the SCGA is used for comparison after the CGA has been run for 2 evolutionary generations since sufficient training data is required to provide useful approximations. Again, similar to the previous experiments, the SCGA identifies improved designs within the same number of fabrications as the CGA, finding a solution yielding $2296rpm$ after 140 fabrications and $2429rpm$ after 160 fabrications (Fig.~\ref{fig:scga-individuals}). Comparing the final 20 array combinations from each experiment, the average rotation speed of the SCGA ($M=2205$, $SD=47$, $N=20$) is significantly greater than the GA-only approach ($M=1894$, $SD=43$, $N=20$) using a two-sample {\it t\/}-test assuming unequal variances, $t(33)=3.7$, $p\le.00076$. Furthermore, the fittest array combination designed by the SCGA ($2429rpm$) was greater than the CGA approach ($2209rpm$) after 160 fabrications.

\section{Conclusions}
This paper has shown that EAs are capable of identifying novel and increasingly efficient VAWT designs wherein a sample of prototypes are fabricated by a 3D printer and examined for utility in the real-world. The use of a neural network surrogate model was found to reduce the number of fabrications required by an EA to attain higher aerodynamic efficiency (rotation speed) of VAWT prototypes. This approach represents the first surrogate-assisted embodied evolutionary algorithm using 3D printing, and completely avoids the use of 3D computer simulations, with their associated processing costs and modelling assumptions. In this case, 3D CFD analysis was avoided, but the approach is equally applicable to other real-world optimisation problems, for example, those requiring computational structural dynamics or computational electro-magnetics simulations. We anticipate that in the future such `design mining' approaches will yield unusual yet highly efficient designs that exploit characteristics of the environment that are extremely difficult to capture in a simulation. In particular, the wind turbine array experiment has shown that it is possible to use surrogate-assisted coevolution to iteratively increase the performance of two closely positioned turbines, taking into account the inter-turbine flow effects, which is especially difficult to achieve under a high-fidelity simulation. The SCGA represents a scalable approach to the design of wind turbine arrays since the number of inputs to the surrogate-models remains constant regardless of the number of turbines undergoing evolution.

For single turbine comparison, the prototype in Fig.~\ref{fig:target-printed}, which is a common 4 blade Savonius, measured 827$rpm$; a $z$-rotated variant similar to the design in U.S. patent 7,371,135 (see Fig.~\ref{fig:patent}) measured 650$rpm$; and a common 3 blade Savonius design (see Fig.~\ref{fig:3blade}) measured 1055$rpm$. That is, the prototypes produced here through artificial evolution are aerodynamically more efficient than several common human designs under the current experimental conditions and performance metrics. 

Although the array experiment did not elicit counter-rotation as might have been expected, evolution is clearly exploiting characteristics unique to its environment. Duplicating the individuals from the fittest array pairing to form homogeneous arrays yields a maximum combined rotation speed of $2359rpm$ compared with $2428rpm$ produced by the heterogeneous array. In addition, while direct comparison cannot be made due to the initial population seeding, the fittest individual from the single turbine $z$-varying experiment (see Fig~\ref{fig:zn-g5-top}) was duplicated to form a homogeneous array and yielded a significantly slower combined rotation speed of $2178rpm$ compared with the $2428rpm$ observed from the array design in Fig~\ref{fig:scga-individuals}.

Future work will include the use of the AC voltage generated by the VAWT prototypes as the fitness computation under various wind tunnel conditions; the coevolution of larger arrays, including the turbine positioning; the exploration of more advanced assisted learning systems to reduce the number of fabrications required; examination of the effect of seeding the population with a given design; investigation of alternative representations that provide more flexible designs including variable number of blades, for example, supershapes~\cite{PreenBull:2014}; and the production of 1:1 scale designs.

The issue of scalability remains an important future area of research. When increasing the scale of designs it is widely known that the changes in dimensionality will greatly affect performance, however it remains to be seen how performance will change in the presence of other significant factors such as turbine wake interactions in the case of arrays. One potential solution is to simply use larger 3D printing and wind-tunnel capabilities whereby larger designs could be produced by the same method. On the opposite end of the spectrum, micro-wind turbines that are $2mm$ in diameter or smaller can be used to generate power, such as for wireless sensors~\cite{Howey:2011}, and in this case more precise 3D printers would be required. Moreover, wind turbines can find useful applications on any scale, e.g., a recent feasibility study~\cite{Park:2012} for powering wireless sensors on cable-stayed bridges examined turbines with a rotor diameter of $138mm$ in wind conditions with an average of $4.4m/s$ (similar to the artificial wind conditions used in this paper.)

If the recent speed and material advances in rapid-prototyping continues, along with the current advancement of evolutionary design, it will soon be feasible to perform a wide-array of automated complex engineering optimisation {\it in situ}, whether on the micro-scale (e.g.,\ drug design), or the macro-scale (e.g.,\ wind turbine design). That is, instead of using mass manufactured designs, EAs will be used to identify bespoke solutions that are manufactured to compensate and exploit the specific characteristics of the environment in which they are deployed, e.g.,\ local wind conditions, nearby obstacles, and local acoustic and visual requirements for wind turbines. 
%
%\bibliographystyle{IEEEtran}
%\bibliography{IEEEabrv,references}

\end{document}